\documentclass[smallcondensed]{svjour3}  
\smartqed 

\usepackage{amsfonts}
\usepackage{amssymb}
\usepackage{mathtools}
\usepackage{pifont}
\usepackage{bbm}
\usepackage{verbatim}

\usepackage{pbox}
\usepackage{cellspace}
\usepackage[keeplastbox]{flushend}
\usepackage{placeins}
\usepackage{dblfloatfix}

\usepackage{hyperref,xspace}

\usepackage{color}
\usepackage{framed}

\usepackage{algorithm} 
\usepackage{algpseudocode}
\usepackage{tabu}

\newcommand{\hl}[1]{{#1}} 

\newcommand{\E}{\mathbb{E}}

\newcommand{\binsketch}{\mathrm{BinSketch}}

\usepackage{paralist,enumitem}

\spnewtheorem*{theoremnonum}{Theorem}{\bf}{\it}
\spnewtheorem{innercustomthm}{Theorem}{\bf}{\it}

\newcommand{\RMSE}{\mathrm{RMSE}}
\newcommand{\tsketch}{{\tt Cabin}\xspace}
\newcommand{\randbin}{{\tt BinEm}\xspace}

\newcommand{\hamest}{{\tt Cham}\xspace}
\newcommand{\tl}[1]{\tilde{#1}}

\begin{document}

\title{Efficient Binary Embedding of Categorical Data  {using BinSketch}
}

\author{Bhisham Dev Verma        \and
        Rameshwar Pratap \and
        Debajyoti Bera~\footnote{Corresponding author}
}


\institute{Bhisham Dev Verma\at
              School of Computing and Electrical Electrical Engineering, \\
              IIT Mandi, H.P., India \\
              \email{d18039@students.iitmandi.ac.in} 
           \and
           Rameshwar Pratap \at
            School of Computing and Electrical Electrical Engineering, \\
              IIT Mandi, H.P., India \\
              \email{rameshwar@iitmandi.ac.in}      
           \and
           Debajyoti Bera \at
           Indraprastha Institute of Information Technology Delhi,\\
           New Delhi, India\\
           \email{dbera@iiitd.ac.in} 
}

\date{Received: date / Accepted: date}

\maketitle
 

\begin{abstract}
  In this work, we present a dimensionality reduction algorithm, {\it aka.} sketching, for categorical datasets. Our proposed sketching algorithm \tsketch constructs low-dimensional binary sketches from high-dimensional categorical vectors, and our distance estimation algorithm \hamest computes a close approximation of the Hamming distance between any two original vectors only from their sketches. 
   The minimum dimension of the sketches required by \hamest to ensure a good estimation theoretically depends only on the sparsity of the data points -- making it useful for many real-life scenarios involving sparse datasets. We present a rigorous theoretical analysis of our approach and supplement it with extensive experiments on several high-dimensional real-world data sets, including one with over a million dimensions. We show that the \tsketch and \hamest duo is a  significantly fast and accurate approach for tasks such as $\mathrm{RMSE}$, all-pair similarity, and clustering when compared to working with the full dataset and other dimensionality reduction techniques.

\end{abstract}


\section{Introduction}
Recent decades have witnessed the ability to generate a large volume of high-dimensional data \hl{arising out of the world-wide-web, IoT, various social media platforms, and applications of finance, biology, etc}. Many of these datasets have large dimensions, sometimes in the order of millions ~\cite{agarwal14a,trillion-data}. A general observation is that many high-dimensional datasets are sparse in nature.  Since computation on such large datasets is cumbersome, requires heavy computational machinery, and often suffers from the ``curse of dimensionality''~\cite{Steinbach2004}, one of the natural approaches to tackle this challenge  is to compute \hl{a low-dimension representation  (\textit{aka sketch}) of each of their vectors which} preserves inherent  geometric properties of the corresponding full-dimensional datasets. 

 In this work, we focus on categorical datasets, where the feature values are from a finite set of categories, e.g., days in a week, colour, month, age group, gender, \textit{etc.} Categorical features appear in many machine learning and data mining applications such as  transaction datasets~\cite{transaction,itemset_categorical,itemset_mining}, 
images~\cite{image_categorical}, in bio-informatics~\cite{DNA,HIV}, in recommendation systems on click-stream data~\cite{click_stream}, {\it etc.}
Further, Hamming distance (HD) appears to be the natural distance metric for categorical data points. Hamming distance between two $n$-dimensional categorical data points $x$ and $y$ is defined as:

\begin{align*}
 HD(x, y)&=\sum_{i=1}^n \mathrm{d}(x_i, y_i) \text{, where }  \mathrm{d}(x_i, y_i)=\begin{cases}
 1,  &\text{if~} x_i\neq y_i, \\
 0,  &\text{otherwise}.
\end{cases}
\end{align*}


A common way to represent categorical data points is via \emph{label-encoding}. If the number of possible categories is $c$, then we use an integer from $\{1,2,\ldots, c\}$ to represent a feature; $0$ is used to represent a missing feature (e.g., if some data points do not have any attribute for ``age group'' while other points have that attribute, we say that the ``age group'' feature is missing in the former points and use 0 for the corresponding feature). In this paper, we focus on developing a sketching algorithm for high-dimensional and sparse categorical vectors. \hl{Sparsity of a feature vector denotes the percentage of zero (missing) entries in it, and the smallest sparsity across all the vectors in a dataset is defined as the sparsity of the dataset. To keep our analysis simple, we define the {\em density} of a label-encoded vector as its Hamming weight which is the number of non-missing features it has. Thus a data point with high sparsity (equivalently, low density) has lots of missing features.}


\paragraph{Problem statement:} Given high-dimensional and \hl{high}-sparsity categorical data points, (i) develop an efficient dimensionality reduction algorithm that compresses the input points into low-dimensional binary vectors, and \hl{(ii) develop an efficient algorithm that can estimate the Hamming distances between the original data points from their sketches}. \\

One of the naive approaches to perform dimensionality reduction for categorical data is to first represent it via binary vectors using {\em one-hot encoding}, where a feature value $x$ is replaced by a $c+1$ dimensional binary vector with $1$ at the position $x$, and $0$ otherwise. We can then further apply known dimensionality reduction algorithms for binary data~\cite{ICDM,oddsketch,JS_BCS,bcs} on those binary vectors. However, this approach becomes impractical when the number of categories is large and may lead to an exponential blowup in the dimension of the resultant binary vector. 

{ A quick browse through tutorials, forums and blogs on the Internet reveals a variety of alternatives~\cite[Tip~3]{10.1371/journal.pcbi.1006907}, \cite{grigorev_2017}, mostly along the lines of the usual dimensionality reduction techniques like PCA, SVD, etc.; however, it is not clear why those techniques, originally developed for real-valued data, should work for discrete-valued data. Some of the techniques require matrix operations, solving an optimisation problem, or running a neural network, all of which are computationally costly. A few techniques are based on hashing or random projections; however, we did not find any  theoretical guarantees they offer towards categorical vectors. It appears that many of the approaches used in practice were not designed for categorical vectors to start with, and have been merely re-purposed, possibly due to the lack of a sound alternative. \hl{ To summarise, the above sketching algorithms are not specifically designed for categorical datasets as they do not offer any provable guarantee on the pairwise Hamming distance estimation from the compressed vectors. To address this gap, in this work we suggest a practical compression algorithm for such datasets that enables the aforesaid estimation with high accuracy. }}

{ Finally, some of the available methods do not produce compressed vectors in binary.}
We specifically want our compressed vectors to be binary for the following reasons. \begin{inparaenum}[(i)]
\item \hl{ Binary vectors are space-efficient as compared to the corresponding real-valued vectors of the same dimension -- one feature of a real-valued vector requires $32$ or $64$ bits of memory (depending on word-size of a CPU), whereas one feature of a binary vector requires only one bit of storage.}
\item \hl{
The binary vectors enable the possibility of using faster bitwise operators during training and inferencing steps which potentially can lead to several benefits such as less memory requirement, lower power consumption, and faster computation of machine learning tasks~\cite{DBLP:conf/icassp/KimS18}.
}
\item \hl{As our sketches are binary, and their pairwise Hamming distances are close to the original pairwise Hamming distance, this enables us to run the same machine learning algorithm on the sketch itself which we may not have been able to run on a categorical dataset due to its high dimensionality.}
\end{inparaenum}

{
The major challenge that we faced is ensuring that the compressed vectors retain information about the pairwise distances between the original points. Even though there are several results known along this direction, e.g., the Johnson Lindenstrauss lemma~\cite{JL83}, Feature Hashing~\cite{WeinbergerDLSA09}, random projections for clustering~\cite{boutsidis2010random}, all of them deal with distances on the Euclidean space. Recently several methods were proposed for distances between discrete vectors, e.g., BCS~\cite{bcs}, however, they were specifically designed for binary vectors without any scope for extension towards categorical vectors.}

\paragraph{Our results:}In this work, we present (i) the \tsketch algorithm that compresses categorical vectors to succinct binary vectors, and (ii) the \hamest algorithm to estimate the Hamming distance between the full-dimension vectors from their reduced-dimension embeddings.

{Compared to the alternatives, our algorithms are designed specifically for categorical vectors, come with sound theoretical guarantees, and have excellent performance in practice. We discuss these benefits below.}

\begin{itemize}
    \item \hl{\textbf{Unsupervised:} 
    Many feature-selection-based compression algorithms for categorical dataset such as $\chi^2$~\cite{chi_square}, mutual information-based feature selection methods~\cite{MI} require labelled data. They essentially compute the correlation between the input features and their respective labels and discard those features whose correlations are smaller than a certain threshold. In contrast, our proposal is completely unsupervised and does not require any labels for compression. }

    \item \textbf{Succinct and sparse embedding:} 
	\hl{\tsketch sketches have an attractive property --- }they are sparse when the original vectors are sparse. 
	In fact, we prove in Lemma~\ref{lem:sparsity} that {the number of ones in the Cabin embedding of a categorical vector $u$ is at most half of the number of non-zero entries in $u$ (in expectation)}. But there is a deeper connection to sparsity. Theoretically speaking, the required dimension of our sketch depends only on the \hl{density of input vectors} and is independent of their original dimension. \hl{Thus, a vector with a large number of dimensions but, at the same time, of high sparsity can be compressed to an extremely small sketch yet retaining its useful theoretical properties. We observed that even smaller sketches sufficed in practice in our empirical exercises.} Moreover, our algorithm outputs a binary-valued sketch which leads to additional space-saving as compared to a real-valued sketch of the same dimension.  Finally, our succinct, sparse, and binary sketches not only provide saving in space required to store the sketch but also enable faster training and inference due to fewer arithmetic computations.

    \item \textbf{Superfast data analysis:} As desired above, our solution generates binary sketches bringing along with it significant advantages against real-valued sketches with respect to space, training, and  inferencing time. Efficiency is further aided by the one-pass nature of \hl{our} algorithms -- \tsketch  and \hamest. Further, the running time complexity of both \tsketch and \hamest is linear in their respective input size. 
	\hl{Empirically, we obtained roughly $136\times$ speedup while generating a similarity heatmap of a Brain-Cell~\cite{genomics20171} dataset with $1.3$ million features, and $112.3\times$ speedup in clustering of a $10^5$-dimensional NYTimes dataset~\cite{UCI} as compared to performing the tasks on the uncompressed full-dimensional datasets.}
%

    \item\textbf{High accuracy:} The biggest advantage of our proposal is that of high-quality sketches. 
   \tsketch is able to compress very-high dimensional points  to low-dimensional sketches such that on several downstream evaluation tasks,  the results on the sketches closely approximate  the corresponding results on the original input. 
 We were able to theoretically explain this behaviour in the form of Theorem~\ref{thm:cham-concentration} which shows that the Hamming distance between two data vectors can be approximated with high accuracy from their sketches if we choose the embedding dimension as $\tilde{O}(s\sqrt{s})$ (ignoring $poly(\log)$ factors over the error probability) where $s$ is an upper bound on the \hl{density} of data. We observe much better compression and accuracy in practice.
    For example, \tsketch is able to compress a  Brain-Cell~\cite{genomics20171} dataset with $1.3$ million features to only $1000$-dimension binary vectors; furthermore, \hamest ensures that we get almost identical looking heat-maps generated on both -- full dimensional and compressed data (see Figure~\ref{fig:CAB_heatmap}). Take another data analytic task, that of clustering. \tsketch is able to compress a $10^5$-dimensional NYTimes dataset~\cite{UCI} to $1000$-dimensions that still generate almost accurate results {\it vis-a-vis} the original dataset. 
	The high accuracy of our proposal is also validated by the variance analysis experiments. We notice that the  variance \hl{of the inaccuracies arising out of \hamest} is small, which reaffirms our claim on accurate estimation of the Hamming distances of original data points (see Figures~\ref{fig:binary_var}, \ref{fig:box_plot_hamming_error}).
Even in our comparison with the related methods, \tsketch is super-fast, yet offering a comparable performance with respect to heat-map generation, \textit{root-mean-square error} ($\RMSE$) test, and clustering. 
\end{itemize}



Our \tsketch and \hamest algorithms follow a two-step approach. In the first step, we reduce the embedding problem over categorical vectors to the same over binary vectors, and in the second step, we solve the embedding problem over binary vectors. The two steps allow us to meet the twin requirements of {\em compressed} and {\em binary} embeddings. The first step is commonly performed in data analysis using one-hot encoding or other deterministic methods. Our proposal is to use a random binary encoding that not only retains the original dimension but also preserves the Hamming distance (see Lemma~\ref{lem:randbin} and Lemma~\ref{lemma:hd_u'_v'}). For the second step, we observed that a few candidates $\binsketch$~\cite{ICDM}, BCS~\cite{bcs}, MinHash~\cite{BroderCFM98}, SimHash~\cite{simhash}, OddSketch~\cite{oddsketch} are already available for embedding binary vectors. We decided to use $\binsketch$ for the second step for a few important reasons. First is that we found $\binsketch$ to perform well in our experiments (see Section~\ref{sec:experiments}). Secondly, some of the alternatives do not have the desired theoretical properties, namely, allow approximation of Hamming distance and generation of binary vectors. $\binsketch$ not only generates binary sketches, but it also allows the approximation of Hamming as well as cosine, Jaccard, and inner product distances, making it a natural choice for us. Lastly, we were able to prove additional properties of $\binsketch$ embeddings that played a crucial role in our main theoretical result which is summarised below.

 {
\paragraph*{\bf Informal version of Theorem~\ref{thm:cham-concentration}:}\quad \it 
Let $\hat{u}, \hat{v} \in \{0, 1\}^d$ denote the $\tsketch$ embeddings of two categorical vectors {$u, v \in \{0, c\}^n$,} respectively, where $c$ denote the number of categories, and $d$ is a suitably chosen constant that depends only on the sparsity of $u$ and $v$. Then the output of $\hamest(\hat{u},\hat{v})$ is close to the Hamming distance of $u$ and $v$,  with high probability.\\
}

 { 
Our objective behind \tsketch and \hamest are two-fold --- state-of-the-art performance in experiments and theoretical bounds to explain the behaviour. Some of the choices we made in the design of our method and empirical evaluation are driven by the second requirement which we believe is crucial for explainable data analysis. In particular, there could be a better alternative to $\binsketch$, and further, an integrated single-step solution for the whole approach. We leave this as a future research direction.} 

\paragraph{Organisation of the paper:}
The rest of the paper is organised as follows. We discuss several related works in Section~\ref{sec:relwork}. In Section~\ref{sec:background}, we briefly revisit the $\binsketch$ algorithm and state other preliminary definitions. In Section~\ref{sec:analysis}, we present our algorithms \tsketch and \hamest and derive their theoretical bounds. In Section~\ref{sec:experiments}, we empirically compare the performance of our proposal on several end tasks with state-of-the-art algorithms. In Section~\ref{sec:conclusion}, we conclude our discussion and state some potential open questions of the work.

\section{Related work}\label{sec:relwork}
Dimensionality reduction or compression of high-dimensional \hl{points} is a well-studied phenomenon in data mining and machine learning. Such algorithms can be broadly grouped into the following four categories: \begin{inparaenum}[(i)] \item random projection, \item feature hashing, \item spectral projection, and \item locality sensitive hashing (LSH). \end{inparaenum} At a high level, all such algorithms compress high dimensional vectors into low-dimensional vectors that maintain some measure of similarity \hl{among} the input vectors. We discuss them below.

	JL-lemma (or random projection)~\cite{JL83} is a seminal algorithm that popularised random projection for dimensionality reduction. Their algorithm essentially projects the input matrix on another matrix (called a projection matrix) \hl{each of whose entries} is sampled from the normal distribution $\mathcal{N}(0, 1)$. The {\em Euclidean distance and inner product of the} low-dimensional vectors obtained after random projection closely approximate the Euclidean distance and inner product, respectively, of the original data points.   {Achlioptas~\cite{DBLP:conf/pods/Achlioptas01} improved over this work by suggesting a faster  algorithm which is particularly suitable in database applications. Their contribution lies in  suggesting a projection matrix whose each entry is  sampled from the \texttt{Rademacher distribution}  ($\{-1, +1\}$ with probability $1/2$),  which leads to faster projection.}  \hl{Realising the need to generate sparse sketches,} Li \textit{et.al.}~\cite{very_sparse_rp} modified random projection by carefully choosing the entries of the projection matrix to make the resultant sketch sparse.

 Similar to random projection, feature hashing~\cite{WeinbergerDLSA09} also offers dimensionality reduction while approximating Euclidean distance and inner product.  It is also known to maintain the sparsity of input in the sketch.  In comparison with \tsketch, it is unclear whether random projection or feature hashing algorithms can be applied on a categorical dataset for approximating Hamming distance. \hl{It can be noted that} similar to feature hashing, \tsketch also maintains the sparsity of the input data in the sketch (see Lemma~\ref{lem:sparsity}).
 

Another recently proposed approach on the lines of random projection is $\binsketch$ ~\cite{ICDM} that suggests an efficient compression algorithm for sparse binary datasets. It compresses high-dimensional binary vectors into low-dimensional binary vectors \hl{and includes a method that accurately} approximates various similarity measures such as Hamming distance, cosine similarity, Jaccard similarity, and inner product. However,  as $\binsketch$ only operates on binary vectors, we can not directly use it for compressing categorical vectors.  One naive approach is to first convert the categorical vectors into binary vectors using \textit{one-hot encoding} and then use $\binsketch$ on it to generate a binary sketch. However, the dimension of a binary embedding obtained via \textit{one-hot encoding}  is $d \times c$, where $d$ and $c$ are the original dimension and the number of categories, respectively. Therefore, this approach becomes impractical when the number of categories and the dimension of input vectors are large. The novelty in our approach is to suggest a {\em randomised} encoding of categorical data into binary vectors of the same dimension that allows us to approximate the original pairwise Hamming distances. $\binsketch$ can be applied to this binary sketch to further compress it into low dimensional binary vectors; the original pairwise Hamming distances can then be approximated from those vectors. Note that there are other known compression algorithms for binary vectors such as BCS~\cite{bcs}. However, we prefer to use $\binsketch$ as it offers both better theoretical as well as practical guarantees on the quality of its estimation. Nevertheless, we included BCS  as a baseline for empirical comparison and observed that our proposal is better in practice.

Principal component analysis (PCA) is another popular method for dimensionality reduction that creates uncorrelated features that successively maximise variance. {Multiple Correspondence Analysis (MCA)}~\cite{MCA} is analogous of PCA but designed for categorical datasets. However, our aim is to estimate the pairwise Hamming distances of the data points from their sketches, whereas that of MCA is to deduce uncorrelated features. 
Locality-sensitive hashing (LSH)~\cite{IM98,simhash,BroderCFM98,GIM99} is another important line of dimensionality reduction algorithms that are primarily used for nearest neighbour search problems. In LSH, points are grouped in a way such that similar points are hashed into the same bucket and dissimilar points hashed into different buckets, with high probability.  Thus, LSH algorithms like SimHash (SH), Hamming-LSH (H-LSH) do not provide explicit estimates of any type of similarities which we aim for in this work. 

There are some learning-based sketching algorithms such as {Latent Semantic Analysis (LSA)}~\cite{LSI}, {Latent Dirichlet Allocation (LDA)}~\cite{LDA}, {Non-negative Matrix Factorisation (NNMF)}~\cite{NNMF},  {Variational auto-encoder (VAE)}~\cite{Kingma2014}, {\it etc.}\ that learn low-dimensional representations while maintaining some inherent properties of the input. \hl{ However, we are not aware of any such technique designed for categorical datasets that maintains or approximates Hamming distances.}
  
\hl{Outside of feature selection and dimensionality reduction,} finding space-efficient sketches for approximating Hamming distance has been well studied in the streaming algorithms framework. Cormode et al.~\cite{CormodeDIM03} described how to compute succinct sketches of large data streams which closely approximates the Hamming distance between the data streams. This result was subsequently improved by Kane et al.~\cite{KaneNW10} by providing an optimal bound on the size of the sketch. These algorithms output real-valued sketches that are optimal only in the asymptotic sense, and are difficult to implement; in contract to those, here we aim for a compression algorithm that outputs binary sketches and are also practically efficient.


\hl{ To the best of our knowledge, there is no dimensionality reduction algorithm available that compresses high-dimensional categorical data into low-dimensional categorical data that can be used to closely approximate the original pairwise Hamming distances. Many sketching algorithms such as PCA, MCA, LSA, LDA, NNMF generate real-valued sketches whereas some others such as SH, H-LSH, Kendall rank correlation test (KT) output discrete sketches but they are not known to approximate pairwise Hamming distances. } 
We nevertheless empirically compare our results with most of the methods mentioned above. We obtained significant speedup in dimensionality reduction time while being able to accurately estimate the Hamming distances from the low-dimensional vectors. We discuss these in detail in Section~\ref{sec:experiments}.

\section{Background}\label{sec:background}
\subsection{Revisiting $\binsketch$ algorithm~\cite{ICDM}}
 {
We first recall $\binsketch$ --- an algorithm to compress sparse binary vectors into binary sketches that preserve several similarity measures such as Hamming distance, inner product, cosine, and Jaccard similarity.}

\begin{definition}[$\binsketch$~\cite{ICDM}]\label{def:binsketch}
  {  Let $\phi$ be a random mapping from $\{1, \ldots n\}$ to $\{1, \ldots k\}$.
    Then a vector $a \in \{0,1\}^n$ is compressed into a vector $a_s \in
    \{0,1\}^{k}$ as $$a_s[j] = \bigvee_{i : \phi(i)=j} a[i],$$
    where $a[i]$ denotes the $i$-th index of the vector $a$, and $\bigvee$ denotes the \texttt{bitwise-OR} operator.}
\end{definition}

 {Theorem $1$ of~\cite{ICDM} (restated below) delivers the guarantee of the estimate of inner product similarity from $\binsketch$ sketches; $\binsketch$ was shown to also approximate a few other similarity measures which are not useful for us.}

\begin{theorem}[Theorem $1$ of~\cite{ICDM} -- Inner product estimation.]
{
Suppose we want to estimate the inner product of two  $d$-dimensional binary
vectors $a, b \in \{0, 1\}^n$, whose \hl{density} is at most $s$, with probability at least $1-\delta$. We
can use $\binsketch$ to construct their  $k$-dimensional binary sketches, where
$k=s \sqrt{{s}/{2} \ln {1}/{\delta} }$. Then using the sketches of $a$ and $b$, the inner product between $a$ and $b$ can be estimated with accuracy $O(\sqrt{s\ln ({1}/{\delta})}).$
}
\end{theorem}

\subsection{Evaluation metrics for clustering performance}\label{subsec:clustering_metrics}
\hl{Consider a clustering task on some dataset.} Let $m$ denote the number of data points,  $\{\omega_1, \omega_2,\ldots, \omega_k\}$ represent the ground-truth clustering results, and $\mathcal{C}=\{c_1, c_2,\ldots, c_k\}$ represent a clustering on the reduced dimension data.  In what follows, we discuss some important metrics to estimate the quality of the clustering $\mathcal{C}$.

\begin{itemize}
    \item \textbf{Purity index:} The \textit{purity index} of $\mathcal{C}$ is defined as
 $$\frac{1}{m}\sum_{i=1}^k \max_{1\leq j\leq k}|\omega_i \cap  c_j|.$$
 The \textit{purity-index} takes a value between $0$ and $1$ --- closer to $1$ indicates better performance. 
 {
\item \textbf{Normalised Mutual Information (NMI)}: The \textit{NMI} of $\mathcal{C}$ is defined as $$\sum_k \sum_j \frac{|\omega_k \cap c_j|}{m} \log \frac{m|\omega_k \cap c_j|}{|\omega_k|\cdot|c_j|}.$$
The \textit{NMI} index takes a value between $0$ and $1$ --- closer to $1$ indicates better performance.
\item \textbf{Adjusted Rand Index (ARI)}:
The \textit{ARI} score of  $\mathcal{C}$ is defined as $$\frac{\sum_{ij} {m_{ij} \choose 2}-\left[\sum_{i} {a_{i} \choose 2} \sum_{j} {b_{j} \choose 2} \right]\big/{m \choose 2} }{{\frac{1}{2} \left[\sum_{i} {a_{i} \choose 2}+ \sum_{j} {b_{j} \choose 2} \right]-\left[\sum_{i} {a_{i} \choose 2} \sum_{j} {b_{j} \choose 2} \right]\big/{m \choose 2}  }}, $$ where $m_{ij}:=|\omega_i \cap  c_j|, a_i:=\sum_j m_{ij},$ and  $b_j:=\sum_i m_{ij}$.
The \textit{ARI} takes a value between $-1$ and $1$ --- closer to $1$ indicates better performance.}
\end{itemize}

\section{\tsketch and \hamest - algorithm and analysis}
\label{sec:analysis}

In this section, we present our \tsketch sketching algorithm and an algorithm to estimate a Hamming distance from the sketches.

Suppose we want to run \tsketch on a dataset with $n$-dimensional vectors and each attribute of a vector could either be missing or must belong to at most $c$ categories (for example, the $5^{th}$ attribute could be day and that can take a value from \{Sunday, Monday, \dots, Saturday\}). We will assume that the categories are represented by $\{1,2,\ldots, c\}$ by some data transformation; if some attribute, say \hl{ the $i$-th one}, is missing then the $i$-th coordinate of the vector will be assigned 0. It is entirely possible to have different sets of categories for each attribute (for example, day of week for the $5^{th}$ attribute and month of year as the $6^{th}$ attribute) as long as we have an upper bound on the largest number of categories of any attribute -- this bound is denoted $c$. So at the end of the data-transformation, we end up with vectors from \hl{$\{0,1,\ldots,c\}^n$ which form the input to \tsketch}.


\tsketch embeds an $n$-dimensional data vector to a $d$-dimensional binary vector where $d \ll n$ and $d$ is chosen as $s\sqrt{\tfrac{s}{2} \ln \tfrac{6}{\delta}}$ in which $s$ denotes an upper bound on the \hl{density} of $u$ and $v$ and $\delta$ is the desired probability of error. It uses two uniformly random mappings,
\begin{enumerate}
    \item \hl{category mapping $\psi: \{0,1,\ldots, c\} {R \atop \to} \{0,1\}$}
    \item \hl{attribute mapping $\pi:\{1,2,\ldots, n\} {R \atop \to} \{1,2,\ldots,d\}$}
\end{enumerate}
and operates in the following two stages.
\begin{enumerate}
    \item Generate an $n$-dimensional binary vector, say $u'$ from an $n$-dimensional category vector, say $u$, using the $\randbin$ algorithm. $\randbin$ uses $\psi$.
    \item Generate a $d$-dimension binary sketch from $u'$. For this step, we chose to use a recently proposed binary sketching technique named $\binsketch$~\cite{ICDM}. $\binsketch$ uses $\pi$. Note that $\binsketch$ can be replaced with any other sketching algorithm for binary vectors that allows us to estimate a Hamming distance with theoretical guarantees.
\end{enumerate}

\begin{figure}
    \centering
    \begin{minipage}[b]{0.5\linewidth}
    \scriptsize
    {\large $\psi$} \begin{tabular}{c|c|c|c|c}
         0 & 1 & 2 & 3 & 4\\
         \hline
         0 & 1 & 0 & 0 & 1 \\
    \end{tabular}\\[1em]
    {\large $\pi$} \begin{tabular}{c|c|c|c|c|c|c}
         1 & 2 & 3 & 4 & 5 & 6 & 7 \\
         \hline
         5 & 6 & 1 & 5 & 6 & 2 & 6 \\
         \hline\hline
         8 & 9 & 10 & 11 & 12 & 13 & 14\\
         \hline
         1 & 3 & 3 & 4 & 4 & 2 & 1 \\
    \end{tabular}
    \end{minipage} %
    \begin{minipage}[b]{0.45\linewidth}
    \includegraphics[width=\linewidth]{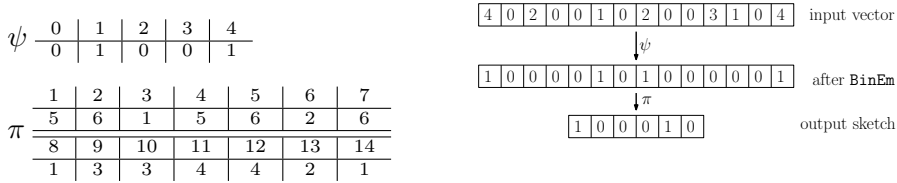}
    \end{minipage} %
    \caption{ {An illustration of \tsketch generating the binary embedding $\langle 100010\rangle$ from a categorical point with the feature vector $\langle 40200102003104\rangle$ (0 values in this indicate missing features). The random mappings that were used are shown on the left.}}
:q
    \label{fig:cabin_algo}
\end{figure}

We describe these two components in Algorithm~\ref{algo:tsketch}; see Figure~\ref{fig:cabin_algo} for an illustration. We include the code for $\binsketch$ for completeness.

    \begin{algorithm}
	\begin{algorithmic}[1]
	    \Function{\tsketch}{$u$} \algorithmiccomment{$u \in \{0,1,\ldots, c\}^n$}
	    \State $u' = \randbin(u)$ \algorithmiccomment{$u' \in \{0,1\}^n$}
	    \State $\tl{u} = \binsketch(u')$ \algorithmiccomment{$\tl{u} \in \{0,1\}^d$}
	    \State \Return $\tl{u}$ 
	    \EndFunction
	    \Function{\randbin}{$u$} \algorithmiccomment{$u \in \{0,1,\ldots, c\}^n$}
	    \State $u'=0^n$
	    \ForAll{non-empty attribute $a$ in $u$}
	    \State let $i$ the position of $a$ \algorithmiccomment{$u_i=a$}
	    \State set $u'_i=\psi(a)$
	    \EndFor
	    \State \Return $u'$
	    \EndFunction
	    \Function{$\binsketch$}{$u'$} \algorithmiccomment{$u' \in \{0,1\}^n$}
	    \State $\tl{u}=0^d$
	    \ForAll{non-zero bit in $u'$}
	    \State let $i$ be the position of the bit
	    \State set $\tl{u}_{\pi(i)}$ to 1
	    \EndFor
	    \State \Return $\tl{u}$
	    \EndFunction
	\end{algorithmic}
	\caption{Sketching algorithm for categorical vectors \label{algo:tsketch}}
    \end{algorithm}

Next, we discuss two \hl{important} properties of the output vectors of \randbin . The first is that these vectors are extremely sparse, and this is the reason we chose $\binsketch$ in stage-2 which is known to be highly efficient for sparse binary data.

\begin{lemma}\label{lem:randbin}
Consider any $u$ with $a$ non-zero attributes. Let $a'$ denote the number of non-zero bits in $\randbin(u)$. Then the following three claims hold. 
$$(a) a' \le a, \quad (b)~\E[a'] = \tfrac{a}{2}, \qquad (c)~\Pr\big[|a' - \tfrac{a}{2}| \ge \epsilon_1 \big] \le \exp{(-\tfrac{2\epsilon_1^2}{a})}.$$ 
\end{lemma}

\begin{proof}
Fact (a) is obvious from the observation that if $u_i=0$ then $u'_i=0$ as well.

For facts (b) and (c), observe that $a'$ can be treated as the sum of $a$ independent and identically distributed binary random variables with $0.5$ success probability; the statements follow from standard analysis of the number of heads among $n$ tosses of a fair coin.
\end{proof}

Let $u'$ denote $\randbin(u)$ and $v'$ denote $\randbin(v)$.
The second property says that the Hamming distance between the outputs of \randbin is sufficient to compute the original distance.

\begin{lemma}\label{lemma:hd_u'_v'} $(a)~HD(u,v) = 2\E[HD(u',v')],$ and $$(b)~ \Pr\big[ |HD(u',v') - HD(u,v)/2| > \epsilon_2 \big] \le \exp(-\tfrac{2\epsilon_2^2}{HD(u,v)}).$$
\end{lemma}

\begin{proof}
For every $i \in \{1,\ldots, n\}$, let $W_i$ be an indicator variable that is equal to 1 when $u_i \not= v_i$, and $W'_i$ be an indicator variable that is equal to 1 when $u'_i \not= v'_i$. Observe that $HD(u,v) = \sum_i W_i$ and $HD(u',v')=\sum_i W'_i$.

We make two observations. First is that $u'_i=v'_i$ whenever $u_i=v_i$. This is true for both the cases of $u_i=v_i=0$ and $u_i=v_i=a \in \{1, \ldots, c\}$. From this observation we get that if $W_i=0$ then $W'_i=0$.

The second observation is that when $u_i \not= v_i$ then $\Pr[u'_i\not=v'_i]=\tfrac{1}{2}$ (or equivalently, $\E[W'_i]=\tfrac{1}{2}$ when $W_i=1$). To see this consider these three cases.
\begin{enumerate}
    \item $u_i=0 \not= v_i$: For this case $u'_i=0$ but $v'_i=1$ with probability $\tfrac{1}{2}$.
    \item $u_i \not=0 = v_i$ (this is similar to the above case)
    \item $u_i \not= 0 \not= v_i$: For this case, $u'_i$ and $v'_i$ are mapped randomly and independently to 0 or 1, so, $u'_i \not= v'_i$ with probability $\tfrac{1}{2}$.
\end{enumerate}
Let $h$ denote $HD(u,v)$. Without loss of generality, we can consider $W_i=1$ for $i=1 \ldots h$, and $W_i=0$ for all other values of $i$. Now from the above observations we can see that $HD(u',v')$ is essentially a sum of $h$ Bernoulli random variables $W'_1, \ldots, W'_h$, each having a probability $\tfrac{1}{2}$ of success. It immediately follows that $\E[HD(u',v')] = \E[\sum_{i=1}^h W'_i] = h/2 = HD(u,v)/2$ proving claim (a). Claim (b) simply says that this sum is tightly concentrated around its mean, and this can be obtained by a straight forward application of Chernoff-Hoeffding's bound for additive error.
\end{proof}

Now we move to our algorithm for estimating Hamming distance between two vectors, say $u$ and $v$ solely from their \tsketch sketches. We call it \hamest and it is described in Algorithm~\ref{algo:hamest}.
It employs an algorithm to estimate the Hamming distance between two binary vectors from their compressions generated by $\binsketch$~\cite[Algorithm~2]{ICDM} --- we refer to this algorithm as {\tt BinHamming} and include it for readability. We have used $D$ to denote $1-\tfrac{1}{d}$ in the algorithm.

    \begin{algorithm}
	\begin{algorithmic}[1]
	    \Function{\hamest}{$\tl{u},\tl{v}$} \algorithmiccomment{$\tl{u},\tl{v} \in \{0,1\}^d$}
	    \State Estimate $\tl{h}=\mathtt{BinHamming}(\tl{u}, \tl{v})$ \algorithmiccomment{\cite[Algo.~2]{ICDM}}
	    \State \Return $2\tl{h}$
	    \EndFunction
	    \Function{{\tt BinHamming}}{$\tl{u},\tl{v}$}
	      \State \hl{Compute $|\tl{u}|$ = Hamming weight of $\tl{u}$}
	    \State \hl{Compute $|\tl{v}|$ = Hamming weight of $\tl{v}$}
	    \State \hl{Compute $\langle \tl{u},\tl{v} \rangle$ = bitwise inner-product of $\tl{u}$ and $\tl{v}$}
	    \State \hl{Compute $\tl{h} = \tfrac{1}{\ln D}\left( D^{|\tl{u}|} + D^{|\tl{v}|} + \tfrac{\langle \tl{u},\tl{v} \rangle}{d} - 1 \right)$}
	    \State \hl{\Return $\tl{h}$}
	    \EndFunction
	\end{algorithmic}
	\caption{Algorithm to estimate Hamming distance\label{algo:hamest}}
    \end{algorithm}

The fact that $\hamest(\tsketch(u), \tsketch(v))$ returns a good estimator of $HD(u,v)$ should be apparent from the results given above and the properties of {\tt BinHamming}. 
We briefly show how to derive a concentration bound on $\tl{h}$ output by {\tt BinHamming} and then formally prove a concentration bound on \hamest with additive accuracy.

\begin{lemma}\label{lem:binhamming}
    Let $h$ denote the Hamming distance of two $d$-dimensional binary vectors $\tl{u}$ and $\tl{v}$, and $\tl{h}$ denote the output of ${\tt BinHamming}(\tl{u}, \tl{v})$. Then, with probability at least $1-\delta$, $|h - \tl{h}| \le 6\sqrt{\tfrac{s}{2} \ln \tfrac{6}{\delta}}$.
\end{lemma}

Lemma~\ref{lem:binhamming} is not proved explicitly \hl{in the $\binsketch$ paper}. However, \hl{it is included as an intermediate step for} proving Lemma~12 (\hl{refer to the upper bound on $W$ in} \cite[Appendix~B]{ICDM}). \hl{We will also require the fact that the \hl{density} (number of ones) of $\tl{u}$ and $\tl{v}$ is at most $s$ which follows from Lemma~\ref{lem:randbin} claim (a).}

Now we come to our main result stating the effectiveness of \hamest. Fix a small number $\delta$ close to zero.
Suppose vectors $u,v \in \{0,1,\ldots, c\}^n$ were compressed first to $u',v' \in \{0,1\}^n$ using \randbin and then to $\tl{u}, \tl{v} \in \{0,1\}^d$ using $\binsketch$. From Lemma~\ref{lem:binhamming} we know that
$$\Pr\Big[ |2 \cdot HD(u',v') - 2\cdot {\tt BinHamming}(\tl{u}, \tl{v})| > 12 \sqrt{\tfrac{s}{2} \ln \tfrac{6}{\delta}} \Big] \le \delta,$$
and using $\epsilon_2=\sqrt{s \ln \tfrac{6}{\delta}}$ in Lemma~\ref{lemma:hd_u'_v'}, we get that
\begin{align*}
    \Pr\Big[ |2 \cdot HD(u',v') - HD(u,v) | > \hl{2}\sqrt{s \ln \tfrac{6}{\delta}} \Big] & \le \exp{(-\tfrac{2s}{HD(u,v)}\ln \tfrac{6}{\delta})}. \\
& \le \exp{(-\ln \tfrac{6}{\delta})} = \tfrac{\delta}{6}.
\end{align*}

For the last inequality, we used the fact that $HD(u,v) \le 2s$ which follows from the sparsity assumption of the input data. Combining these two inequalities, we arrive at
\hl{
\begin{align*}
    & & \Pr\Big[ & |2HD(u',v') - HD(u,v)| > 2\sqrt{s \ln \tfrac{6}{\delta}} \mbox{ \quad OR } \\
    & &	    & |2HD(u',v') - 2\cdot {\tt BinHamming}(\tl{u}, \tl{v})| > 12 \sqrt{\tfrac{s}{2} \ln \tfrac{6}{\delta}} \Big] \le \tfrac{\delta}{6} + \delta = \tfrac{7\delta}{6}\\
    & \implies & \Pr\Big[ & |2HD(u',v') - HD(u,v)| \le 2\sqrt{s \ln \tfrac{6}{\delta}} \mbox{ \quad AND } \\
    & &	    & |2HD(u',v') - 2\cdot {\tt BinHamming}(\tl{u}, \tl{v})| \le 12 \sqrt{\tfrac{s}{2} \ln \tfrac{6}{\delta}} \Big] \ge 1- \tfrac{7\delta}{6} \\
    & \implies & \Pr\Big[ & |HD(u,v) - 2\cdot {\tt BinHamming}(\tl{u}, \tl{v})| \le (2 + \tfrac{12}{\sqrt{2}})\sqrt{s \ln \tfrac{6}{\delta}} \Big] \ge 1 - \tfrac{7\delta}{6}\\
    & \implies & \Pr\Big[ & |HD(u,v) - 2\cdot {\tt BinHamming}(\tl{u}, \tl{v})| \le 11\sqrt{s \ln \tfrac{6}{\delta}} \Big] \ge 1 - \tfrac{7\delta}{6}\\
    & \implies & \Pr\Big[ & |HD(u,v) - 2\cdot {\tt BinHamming}(\tl{u}, \tl{v})| \ge 11\sqrt{s \ln \tfrac{6}{\delta}} \Big] \le \tfrac{7\delta}{6},
\end{align*}
and adjusting the probabilities further, we obtain our main result for $\hamest$.
}

\begin{theorem}\label{thm:cham-concentration}
    For a small $\delta \in (0,1)$, and $n$-dimensional category vectors $u,v$, $|\hamest(\tl{u},\tl{v}) - HD(u,v)| > 11\sqrt{s \ln \tfrac{7}{\delta}}$ with probability at most \hl{$\delta$}, where $\tl{u}$ and $\tl{v}$ are the outputs of $\tsketch(u,v)$, respectively.
\end{theorem}

Even though the theorem requires that the \hl{density} of $u,v$ be at most $s$ and then \hl{stipulates that} the estimate of the Hamming distance has at most $O(\sqrt{s})$ additive error, we should point out that these bounds are obtained using loose probability inequalities. We show in the next section that the performance is superior on real datasets.

We end this section with a quick fact that \tsketch retains, in fact, improves sparsity. We prove this fact below in expectation.

\begin{lemma}\label{lem:sparsity}
Consider any $u \in \{0,1,\ldots, c\}^n$ with $T$ non-zero values, and let $\tl{u}$ denote $\tsketch(u)$ with $\tl{T}$ non-zero values. Then, $\E[\tl{T}] \le T/2$.
\end{lemma}

\begin{proof}
Let $\randbin(u)$ be denoted $u'$ and let $T'$ be a random variable denoting the number of ones in $u$. We proved in Lemma~\ref{lem:randbin} that $\E[T']=\tfrac{T}{2}$. Now, $\E[\tl{T}]$ can be expressed as the number of non-empty bins when $T'$ balls are thrown uniformly into $d$ bins; we obtain $\E[\tl{T}|T'] = d-d(1-\tfrac{1}{d})^{T'}$. Therefore, $\E[\tl{T}] = \E[\E[\tl{T}|T']] = d - d\cdot \E[(1-\tfrac{1}{d})^{T'}]$. We can bound $\E[(1-\tfrac{1}{d})^{T'}] \ge (1-\tfrac{1}{d})^{\E[T']}$ using Jensen's inequality, and using Lemma~\ref{lem:randbin} and a standard binomial inequality, prove that $\E[\tl{T}] \le d - d \cdot (1-\tfrac{1}{d})^{T/2} \le \tfrac{T}{2}$.
\end{proof}

 \noindent\textbf{Computational complexity of \tsketch and \hamest:}
The $\binsketch$ and $\randbin$ \hl{subroutines} stated in \tsketch (Algorithm~\ref{algo:tsketch}) are one-pass \hl{methods}. The complexity of $\binsketch$ is linear in the number of data points and the data dimension. The complexity of $\randbin$ is  linear in the number of data points and the sketch dimension. Note that the dimensionality of the sketch is $O(s^{3/2})$, which is independent of the original dimension of data, where $s$ is the \hl{density} of the input.  Therefore, the overall running time complexity of  \tsketch  remains linear in the number of data points and the data dimension.

Further, \hamest (Algorithm~\ref{algo:hamest}) \hl{makes} one pass over the sketch obtained via Algorithm~\ref{algo:tsketch} to  estimate a Hamming distance. Therefore the complexity of \hamest is linear in the number of data points and the  sketch dimension. 

\section{Experiments}\label{sec:experiments}

\noindent\textbf{Hardware description:} We performed our experiments on a Tyrone DS300TR-34 machine having the following configuration: CPU: Intel(R) Xeon(R) E5-2630 v4 CPU @ 2.20GHz x 20, 32 GB RAM, Windows 10 Professional OS.

\begin{table}[!t]
%
\centering
  \caption{ Datasets used for empirical study, ordered according to dimension.}
  \addtolength{\tabcolsep}{-3pt}
  \label{tab:datasets}
  \begin{tabular}[t]{lrrrrr}
  \hline
      Datasets   &  Categories    & Dimension   &   \hl{Sparsity} & \hl{Density} & Number of points\\
  \hline
      \raggedright\textrm {KOS blog entries}~\cite{UCI}                 & $42$	& $6,906$	& $93.38\%$ & 457 & 3,430 \\
      \raggedright\textrm {NIPS full papers}~\cite{UCI}			& $132$ & {$12419$}	& $92.64\%$ & 914 & 1,500 \\
      \textrm{Enron Emails}~\cite{UCI}                                  & $150$	& $28,102$	& $92.81\%$ & 2,021 & 39,861 \\
      \raggedright\textrm{NYTimes articles}\cite{UCI}                   & $114$	& $102,660$	& $99.15\%$ & 871 & 10,000 \\
      \raggedright\textrm{PubMed abstracts}\cite{UCI}                   & $47$	& $141,043$	& $99.86\%$ & 199 & 10,000 \\
      \textrm {Million Brain Cells, E18 Mice}\cite{genomics20171}	& $2,036$ & $1,306,127$	& $99.92\%$ & 1,051 & 2,000 \\
  \hline
\end{tabular}%
\addtolength{\tabcolsep}{3pt}
\end{table}

\noindent\textbf{Datasets:} The efficacy of our proposal is best described for high-dimensional datasets. The \hl{categorical datasets that we found to be publicly available were} mostly low-dimensional, therefore we considered several integer-valued freely available real-world datasets as categorical. We chose five datasets (see Table~\ref{tab:datasets}) for our experiments. The dimensions of the data in these datasets ranged from $6906$ to $1.3$ million, their sparsity ranged from $92.64\%$ to $99.92\%$, and the number of categories varied from $42$ to $2036$. We observed that on certain experiments/datasets, many baselines suffered from \textit{``out-of-memory” (OOM)} and \textit{``did not stop” (DNS)} even after running them for a sufficiently long time. Therefore, for those instances we \hl{conducted the experiments on a randomly sampled subset of the corresponding} dataset. 
\hl{The datasets are of two types.}

\begin{itemize}[wide,nosep]
    \item \texttt{BoW (Bag-of-words)}~\cite{UCI}: We consider the following \hl{five} datasets -- KOS blog entries,  {NIPS full papers}, Enron Emails,  NYTimes news articles, and PubMed abstracts --  that have ``BoW'' (bag-of-words) representations of the corresponding text corpora. In all these datasets, the attributes \hl{represent} the frequency of the words appearing in the documents. Since these frequencies take integer values,  we consider them as categories.  
    \item \texttt{1.3 Million Brain Cell Dataset}~\cite{genomics20171}: This dataset consists of the results of single-cell RNA-sequencing \texttt{(scRNA-seq)} of $1.3$ million cells captured and sequenced from an \texttt{E18.5} mouse brain, and made available in public by 10x Genomics\cite{genomics20171} ~\footnote{\url{https://support.10xgenomics.com/single-cell-gene-expression/datasets/1.3.0/1M_neurons}}. Each gene represents a data point and for every gene, the dataset stores the \hl{integer-valued} read-count of that gene corresponding to each cell -- these read-counts form our features. 
 \end{itemize}
 
 \begin{table}[!ht]
\centering
  \caption{ Baseline algorithms.} 
  \label{tab:baselines}
\begin{tabular}[t]{l}
\hline
\textrm{Binary Compression Scheme (BCS)}~\cite{bcs}~*\\
\textrm{Hamming LSH (H-LSH)}~\cite{GIM99}~*\\
\textrm{Feature Hashing} (FH)~\cite{WeinbergerDLSA09}\\
\textrm{Signed-random projection/SimHash} (SH)\cite{simhash}\\
\textrm{Kendall rank correlation coefficient (KT)}\cite{kendall1938measure}\\
\textrm{Latent Semantic Analysis (LSA)}\cite{LSI}\\
\textrm{Latent Dirichlet Allocation (LDA)}\cite{LDA}\\
\textrm{Multiple Correspondence Analysis (MCA)}\cite{MCA}\\
\textrm{Non-neg. Matrix Factorisation (NNMF)}\cite{NNMF}\\
\textrm{ {Variational auto-encoder (VAE)}}~\cite{Kingma2014}\\
vanilla \textrm{Principal component analysis (PCA)}\\
\hline
\footnotesize{* BCS and H-LSH are applied on a \randbin embedding}\\
\end{tabular}
\end{table}

\noindent\textbf{Baseline algorithms:} 
The alternative approaches that we compare against are listed in Table~\ref{tab:baselines}. 
To the best of our knowledge, there is no off-the-shelf unsupervised dimensionality reduction method available which gives low-dimensional binary embedding of a categorical dataset while approximating Hamming distance. Therefore, we identified some of the state-of-the-art unsupervised dimensionality reduction algorithms for empirically evaluating our solution.
\hl{Our method is completely unsupervised and doesn’t require labels of the data points for dimensionality reduction. Nevertheless, to evaluate quality of data-analytics tasks, we also included some supervised feature selection methods such as $\chi^2$~\cite{chi_square} and mutual information-based~\cite{MI} which compute the correlation of features and labels before performing feature selection.}

Recall that our method works in a two-step manner: \begin{inparaenum}[(i)]
\item we first compute the binary embedding of the given categorical data points, and,  
\item then we further compress binary vectors obtained in the first step using \hl{$\binsketch$}~\cite{ICDM}.  \end{inparaenum}
    The $\binsketch$ algorithm due to Pratap \textit{et.al.}~\cite{ICDM} is a state-of-the-art algorithm for computing low-dimensional binary vectors for given high-dimensional binary vectors while simultaneously approximating Hamming distance, inner product, Jaccard, and cosine similarity in the same sketch. It is also possible to use Hamming-LSH~\cite{GIM99} and BCS~\cite{bcs} for the second step instead of $\binsketch$, i.e., after generating a binary vector using \randbin. We treat these combinations of \randbin+Hamming-LSH \hl{(denoted H-LSH)} and \randbin+BCS \hl{(denoted BCS)} as two other possible techniques.\\


 \noindent\textbf{Reproducibility details:}  We implemented {Feature Hashing (FH)}~\cite{WeinbergerDLSA09},   {SimHash (SH)}\cite{simhash}, BCS~\cite{bcs}, Hamming-LSH~\cite{GIM99}, and BinSketch~\cite{ICDM} algorithms on our own. Hamming-LSH is implemented by randomly sampling $d$ features \hl{($d$ denotes the embedding dimension)} from each data point, computing the Hamming distance restricted to the sampled features, and then scaling it appropriately for the full dimension. We made all these implementations freely available~\footnote{\scriptsize{\url{https://github.com/Vicky175/Cabin_Cham}}}. 
  We used \texttt{pandas DataFrame}\footnote{\scriptsize{\url{https://pandas.pydata.org/pandas-docs/stable/reference/api/pandas.DataFrame.corr.html}}} for Kendall rank correlation coefficient~\cite{kendall1938measure}.
 For Latent Semantic Analysis (LSA)~\cite{LSI}, Latent Dirichlet Allocation (LDA)~\cite{LDA}, Non-negative Matrix Factorisation (NNMF)~\cite{NNMF}, Variational auto-encoder (VAE)~\cite{Kingma2014}, and  vanilla Principal component analysis (PCA), we used their implementations available from the \texttt{sklearn.decomposition} library \footnote{\scriptsize{\url{https://scikit-learn.org/stable/modules/classes.html\#module-sklearn.decomposition}}}.
 For Multiple Correspondence Analysis (MCA)~\cite{MCA}, we used an existing Python implementation~\footnote{\scriptsize{\url{https://pypi.org/project/mca/}}}.

We used \texttt{numpy} arrays for storing our vectors. We  invoked  {\texttt{numpy.sum($u != v$)}} for computing the Hamming distance and \texttt{numpy.dot($u, v$)} for computing the inner product between two data points $u$ and $v$.
 
\begin{figure*}[ht]
\centering
\includegraphics[width=0.9\linewidth]{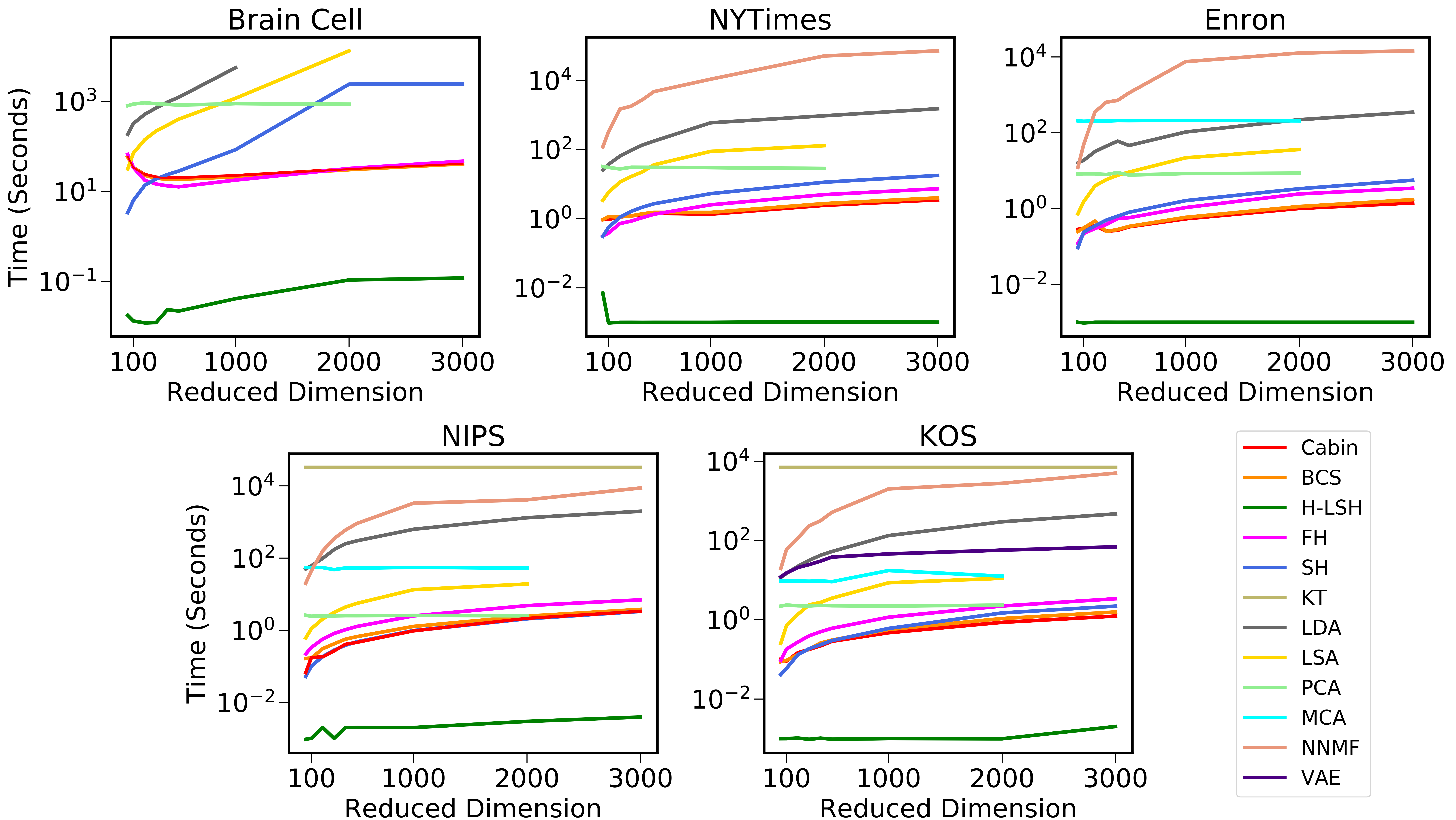}
\caption{Comparing the speed of dimensionality reduction. PCA, MCA, and LSA cannot compress beyond the number of data points and the original dimension, leading to missing values beyond a certain point. LDA did not stop after $20$ hours for the reduced dimensions more than $1000$. { Note that many other baseline methods also did not finish or ran out of memory as the dimensions of the datasets increased (also refer to Table~\ref{tab:speed_up_dim_time}). Among those which finished successfully, Cabin is only slower than Hamming-LSH, but the latter displays a worse performance in the RMSE and the clustering experiments (see Figures \ref{fig:rmse}, \ref{fig:purity}, \ref{fig:nmi}, and \ref{fig:ari}).}}
\label{fig:dim_reduction_time}
\end{figure*}

{\tabulinesep=5cm
\begin{table*}[t]
\centering
    \caption{{Speedup of \tsketch ~ \textit{w.r.t.} baselines on the reduced dimension $1000$. 
} }\label{tab:speed_up_dim_time}
\scalebox{0.75}{
  \begin{tabular}{lccccccccccc}
    \hline
     Dataset    &   NNMF          &  MCA    &  {VAE}        &      LDA        &     LSA         &    PCA         &      FH       &      SH       &   KT    &    BCS       & H-LSH\\
\hline
Brain Cell & \texttt{DNS}    & \texttt{OOM} &  {\texttt{OOM}}  & $251.76 \times$ & $52.77 \times$  & $40.08\times$  & $0.803\times$ & $3.80\times$  & \texttt{OOM}      & $0.95\times$  & $0.0018\times$\\
PubMed     & $11250.6 \times$       & \texttt{OOM}&  {\texttt{OOM}}  & $131.83\times$ & $30.42  \times$  & $24.82\times$  & $0.752\times$ & $4.99\times$  & \texttt{OOM}      & $1.05\times$  & $0.0009\times$\\
NYTimes    & $8061 \times$   & \texttt{OOM}  &  {\texttt{OOM}} & $441.44 \times$ & $65.56 \times$  & $22.2\times$   & $1.864\times$ & $3.92\times$  & \texttt{OOM}      & $1.12\times$  & $0.0007\times$\\
Enron      & $9956.8 \times$ & $392.76 \times$&  {\texttt{OOM}}& $195.46 \times$ & $40.80\times$   & $15.67\times$  & $1.987\times$ & $3.02\times$  & \texttt{DNS}      & $1.09\times$  & $0.0019\times$\\
 {NIPS}       &  {$3477.9\times$} &  {$ 54.61\times $}&  {\texttt{OOM}} &  {$358.4\times$} &  {$13.94\times$}   &  {$5.84\times$ }  &  {$2.51\times$} &  {$1.02\times$}  &   {$34493\times$}   &  {$1.34\times$}  &  {$0.0019 \times$}\\
KOS        & $4255.45\times$ & $37.06 \times $& $91.2 \times $& $282.33 \times$ & $18.24\times$   & $4.72\times$   & $2.467\times$ & $1.28\times$  &  $14932 \times$   & $1.20\times$  & $0.0021\times$\\
\hline
 \end{tabular}}
     \footnotesize{\texttt{OOM} indicates ``out-of-memory error'' and {\tt DNS} indicates ``did not stop'' even after 20 hours.}
\end{table*}
}

\subsection{Speed of dimensionality reduction}\label{subsubsec:dim_red_time}

    A comparison of the time taken for dimensionality reduction on the datasets mentioned above is illustrated in Figure~\ref{fig:dim_reduction_time} and Table~\ref{tab:speed_up_dim_time}. We notice that Hamming LSH is the fastest, however, its performance in the $\RMSE$ experiments (Subsection~\ref{subsubsec:rmse}, Figure~\ref{fig:rmse}) and the other tasks (Subsection~\ref{subsec:end_task_comparision}, Figure~\ref{fig:purity}) is significantly worse. The speed of \tsketch is comparable with that of Feature Hashing, SimHash, and BCS, however, the \hl{latter} too suffer from an inaccurate estimation of Hamming distance leading to fairly poor performance according to the $\RMSE$ measure (Subsection~\ref{subsubsec:rmse}) and the other tasks (Subsection~\ref{subsec:end_task_comparision}).
  We want to draw attention to the fact that many baseline methods such as VAE, MCA, KT give \textit{out-of-memory (OOM)} error or their reported running time is quite high, especially on high values of input or output dimensions. Therefore, we could not perform dimensionality reduction for all dimensions with some of the algorithms.


\subsection{Quality of sketches using root mean square error}\label{subsubsec:rmse}
{}
We evaluate the quality of sketches obtained using different approaches by comparing the error in Hamming distance estimation. For this, we define the Hamming error for a pair of points $u,v$ as 
    \begin{align*}
	HE(u,v)  = & \mbox{actual Hamming distance between $u$ and $v$} \\
		 & - \mbox{\hl{estimated Hamming distance obtained from their sketches}}.
    \end{align*}
    To evaluate the quality of \tsketch sketches, we compare its {root-mean-squared-error ($\RMSE$)} with \hl{the relevant baseline algorithms}. $\RMSE$ is defined as $\sqrt{\sum_{u,v} HE(u,v)^2/N}$ where $N$ represents the total number of pairs. It is a standard metric for comparing dimensionality reduction algorithms and a lower value indicates better performance.
For this experiment we compare our solution with BCS, Hamming LSH, Kendall rank correlation coefficient, Feature Hashing, and  SimHash. Feature Hashing and SimHash are known to approximate inner product and cosine similarity, respectively, which do not have a direct relation to Hamming distance; however, we include them in the comparison nonetheless since they output discrete sketches, and so, Hamming distance can be defined on them. We did not find it meaningful to compare with the methods that output real-valued sketches.

\begin{figure*}[ht]
\centering
\includegraphics[width=0.9\linewidth]{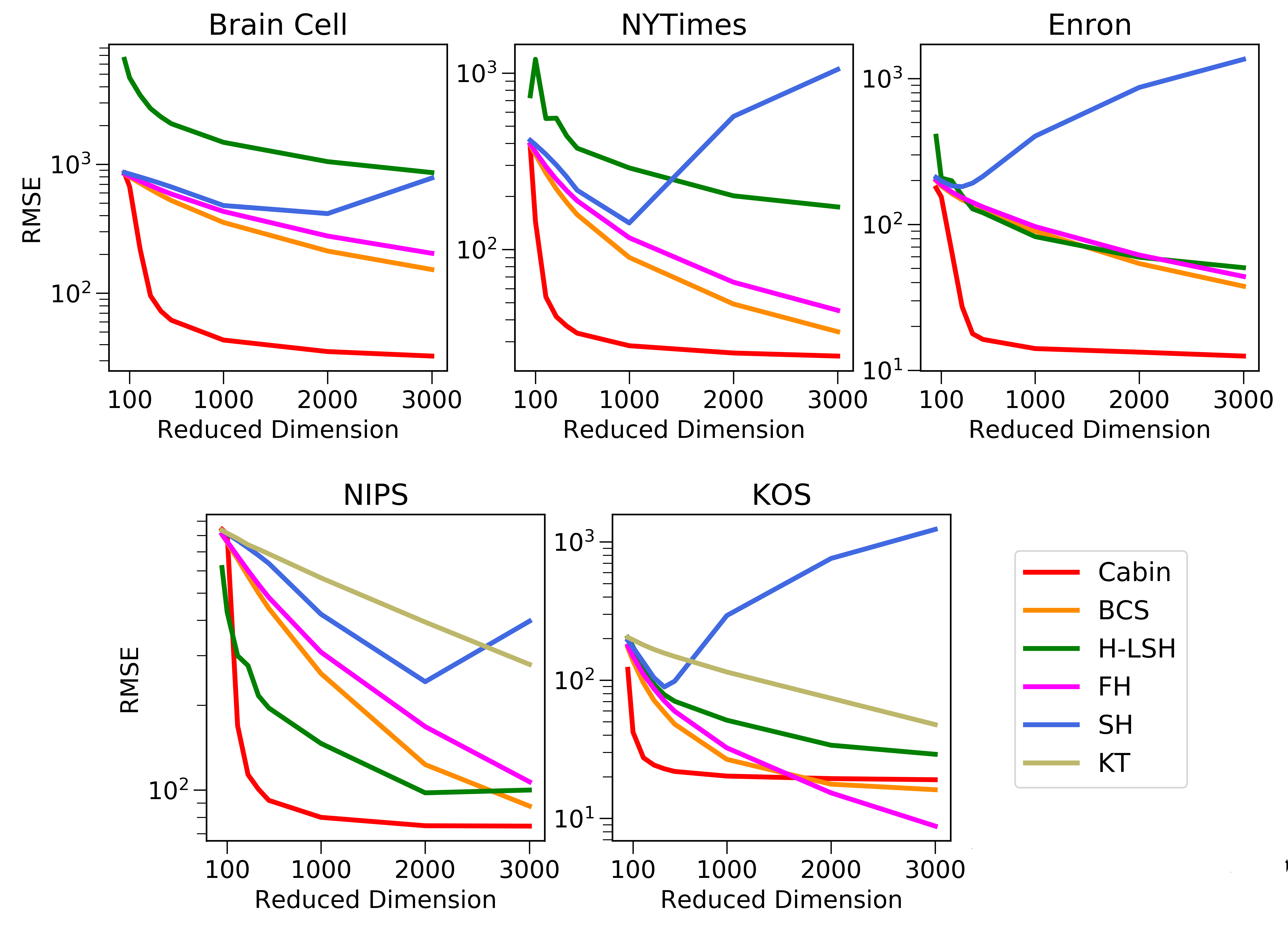}
\caption{{Comparison on $\RMSE$ { (root-mean-squared-error)} among baselines. A lower value is an indication of better performance. On Enron dataset KT couldn’t finish in $10$ hours, and on Brain cell and NYTimes datasets, it gave  an out-of-memory  error.
}}
\label{fig:rmse}
\end{figure*}

%
%

\hl{We performed the RMSE experiment on a sample of $2000$ data points from each dataset. Note that this experiment requires generating all pairwise distances, which in our case amounts to ${2000 \choose 2}\approx 1,999,000$ pairs. Therefore including more than $2000$ points would be tedious.}
The results of $\RMSE$ comparison are illustrated in Figure~\ref{fig:rmse} where we 
notice that the $\RMSE$ of \tsketch is the lowest, and rapidly reaches a very low value at reduced dimensions of a few hundred. This further indicates that \tsketch can compress to much smaller dimensions with a negligible loss in quality compared to the other discrete-valued sketches.%

We wish to point out an interesting trend for some of the hashing-based methods, namely, Cabin, Feature Hashing (FH), and BCS. They tend to perform better when there are few hash collisions, and this is pronounced when the embedding dimension is large and the input vectors are sparse. Indeed, all three show a remarkable trend of low RMSE as their embedding dimensions are increased for all the datasets. This effect gains prominence for KOS whose dimension is less than 7000 and whose sparsity is about $93.4\%$. Thus, as the embedding dimension is increased to 2000 (a very high value compared to the original dimension), BCS and FH outperform Cabin; however, the latter still remains the better choice for KOS when the embedding dimension is less than 1000.


\subsection{Analysis of \tsketch sketches}
Recall that \tsketch~ generates low-dimensional binary embeddings in a two-step process: \begin{inparaenum}[(i)]
\item \randbin first computes the binary embedding of a given categorical vector, and then,
\item $\binsketch$ compresses that binary vector~\cite{ICDM}.  \end{inparaenum}
Intrigued by the low error in the Hamming distances estimated from the \tsketch sketches, we decided to dive deeper and check the accuracy of both these steps.
 
\begin{figure*}[ht]
\centering
\includegraphics[width=0.6\linewidth]{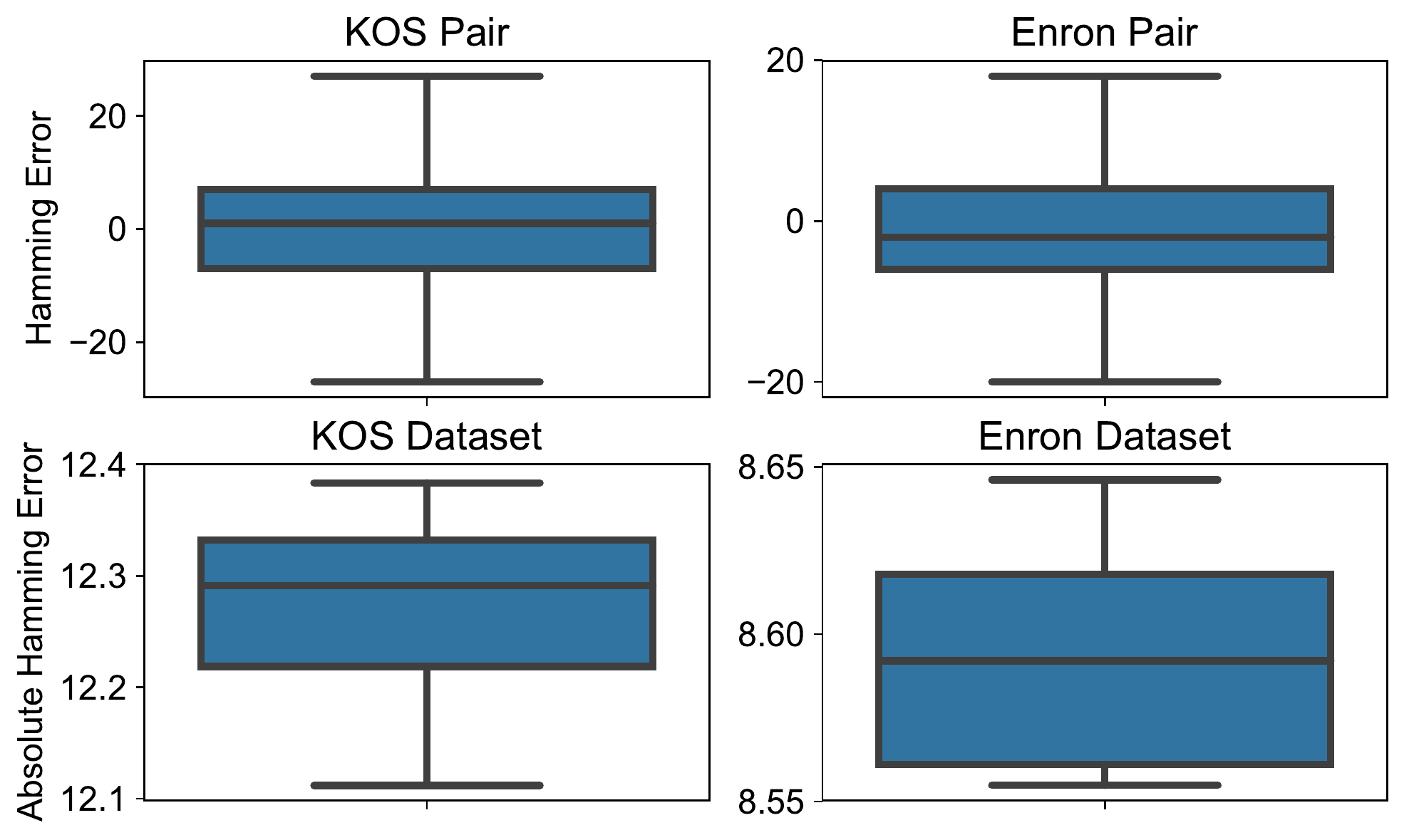}
\caption{Variance analysis of \randbin embeddings. {The pair of box-plots in the  first row  displays the  Hamming errors for a randomly sampled pair of data points, whereas the bottom pair 
 of box-plots show the average of the absolute Hamming errors for all pairs of points.}
 }
\label{fig:binary_var}
\end{figure*}

\subsubsection{Analysis of the first step: \randbin}
We conducted two experiments to understand the efficacy of \randbin. In the first experiment we chose two random data points (say, $u,v$), obtained the difference between $HD(u,v)$ and $HD(\randbin(u), \randbin(v))$ and generated a box-plot of these errors obtained from 1000 independent trials for the same $u,v$, by generating random binary embedding via $\randbin$. The purpose was to understand how much the random embeddings overestimate or underestimate a particular Hamming distance. It is evident from the first two plots of Figure~\ref{fig:binary_var} that \randbin embeddings preserve the actual Hamming distance pretty accurately and appears to be almost uniformly distributed around the actual value.

In the second experiment, we box-plot the average Hamming error $$\sum_{u,v} |HD(u,v) - HD(\randbin(u), \randbin(v))|/N,$$ obtained from 1000 independent runs of \randbin (here $N$ denotes the total number of pairs). We added only the absolute errors since the errors could be seen to be both positive and negative. The last two plots of Figure~\ref{fig:binary_var} make it clear that \randbin embeddings are highly consistent, with very little variance, and with a low average error. These two experiments explain that there is very little loss in the Hamming distance information in the first step of \tsketch.

\begin{figure*}[ht]
\centering
\includegraphics[ width=\linewidth]{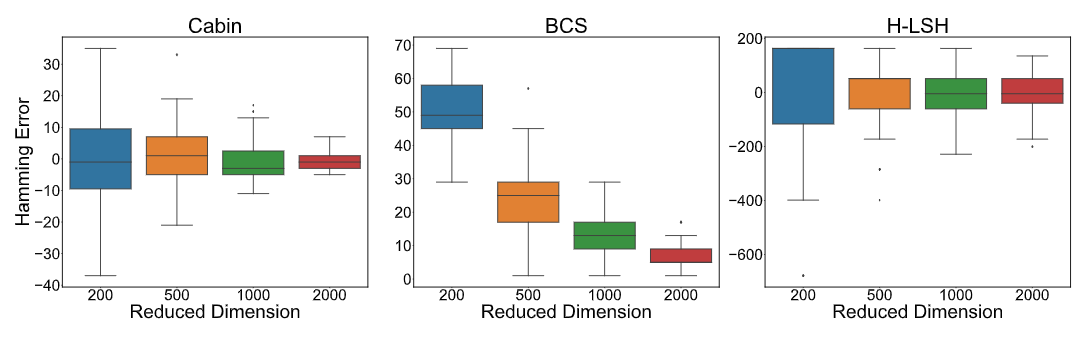}
\includegraphics[ width=0.66\linewidth]{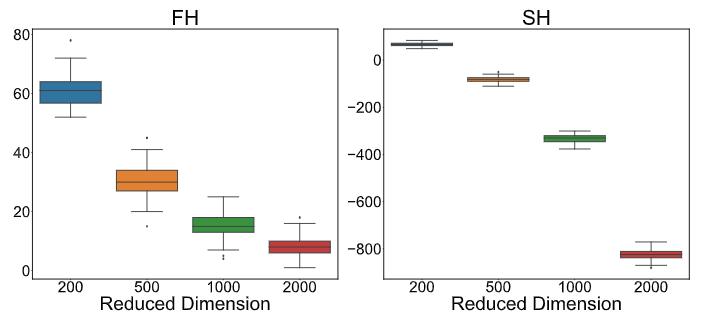}
    \caption{Analysis of Hamming error in compressing the \randbin embeddings for a random pair from the Enron dataset.}
    \label{fig:box_plot_hamming_error}
\end{figure*}

\subsubsection{Analysis of the second step: $\binsketch$} 
Here we conducted an experiment to ask the question: Why $\binsketch$? Recall that we gave a theoretical justification of this choice in Section~\ref{sec:analysis} that we now wanted to supplement. For the experiment, we chose a random pair of points (say $u,v$) from the Enron dataset, generated their binary embeddings $u',v'$ via \randbin, and then compressed them using $\binsketch$ and other discrete sketching methods -- BCS, Hamming-LSH, Feature Hashing, and SimHash, to various reduced dimensions. As before, we computed the Hamming error between the pairs, repeated this process $1000$ times, and generated the corresponding box-plots which we present in Figure~\ref{fig:box_plot_hamming_error}.
%
%
We observe that the expected value of the Hamming error is close to zero when $\binsketch$ is employed, for all the dimensions. For BCS and Feature Hashing, the expected error is large at low dimensions. The behaviour of Hamming-LSH is close to that of \tsketch with a slightly worse variance. In fact, the variance of the sketches obtained using $\binsketch$ is the lowest among the alternatives, and moreover, the variance starts decreasing rapidly as the reduced dimension is increased.
 
So with these experiments, we have empirically validated that the Hamming distances estimates are close to the actual distance on an average, and further, have a low variance. That explains why \hamest is able to estimate the pairwise Hamming distances from the \tsketch sketches with high precision.


\subsection{Performance of clustering}\label{subsec:end_task_comparision}
Next, we conducted experiments to understand if the \tsketch sketches are suitable for data analytic end tasks. The first task we chose was {\em clustering}, and the aim was to verify { if those sketches can compete with the compressed vectors obtained from the standard dimensionality reduction approaches, when used for clustering.}
\hl{For the clustering experiments on NYTimes and PubMed, we used a random sample of 10,000 points since several baseline algorithms started throwing \textit{OOM/DNS} on more points or on the entire dataset. KOS, Enron and NIPS datasets were considered in their entirety.}
We do not include clustering results on the Brain Cell dataset since the clustering process faced an ``out-of-memory'' error on the full-dimensional dataset.

\begin{figure*}[ht]
\centering
    \includegraphics[width=0.8\linewidth]{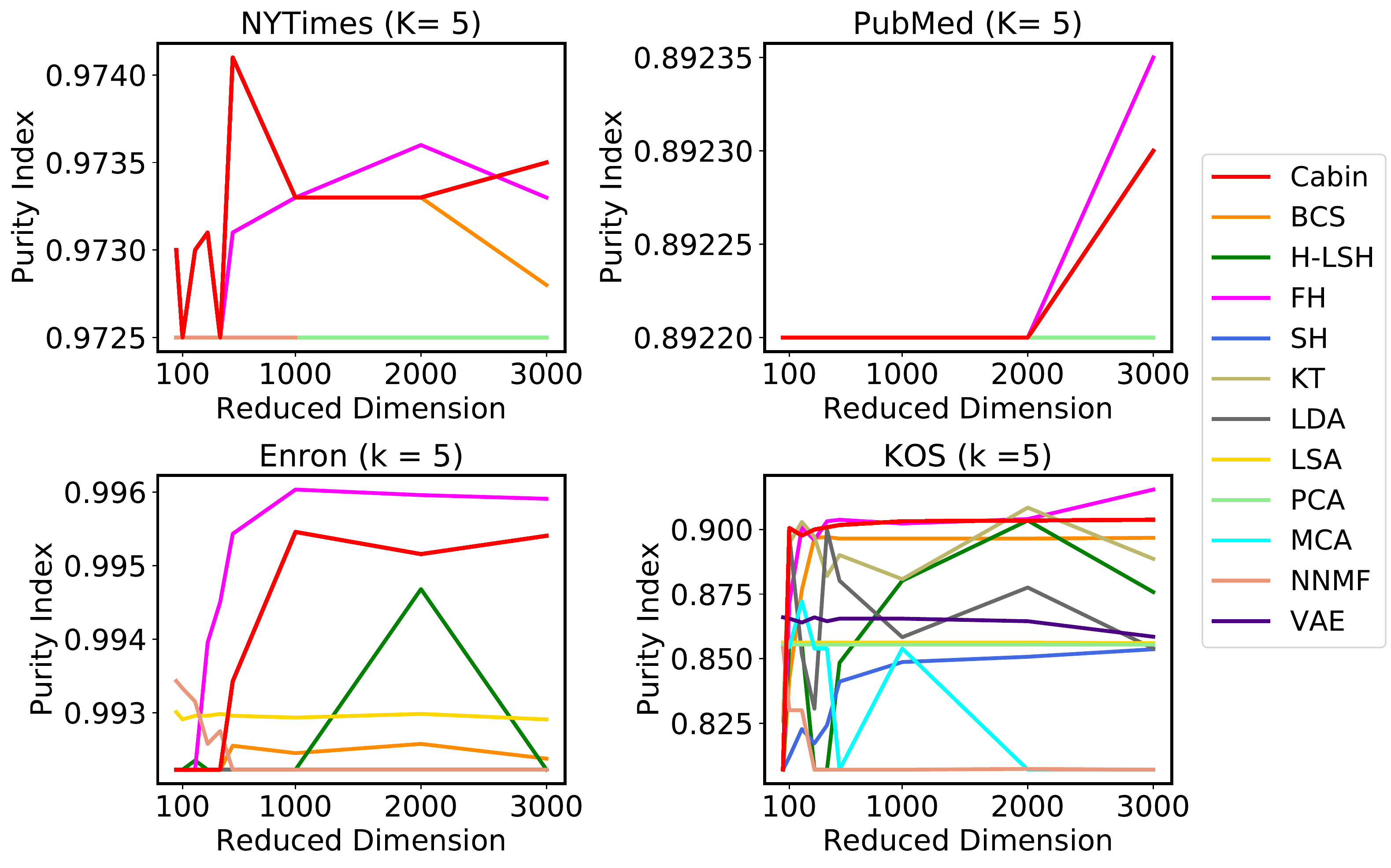}
\caption{\hl{Comparing the quality of clusters on the compressed datasets using \textit{purity-index} metric. The purity-index takes a value between $0$ and $1$. A higher value indicates a better performance. {Observe that the performance of Cabin is among the top few at 1000 or more embedding dimensions.}
}}
\label{fig:purity}
\end{figure*}

\begin{figure*}[ht]
\centering
    \includegraphics[width=0.8\linewidth]{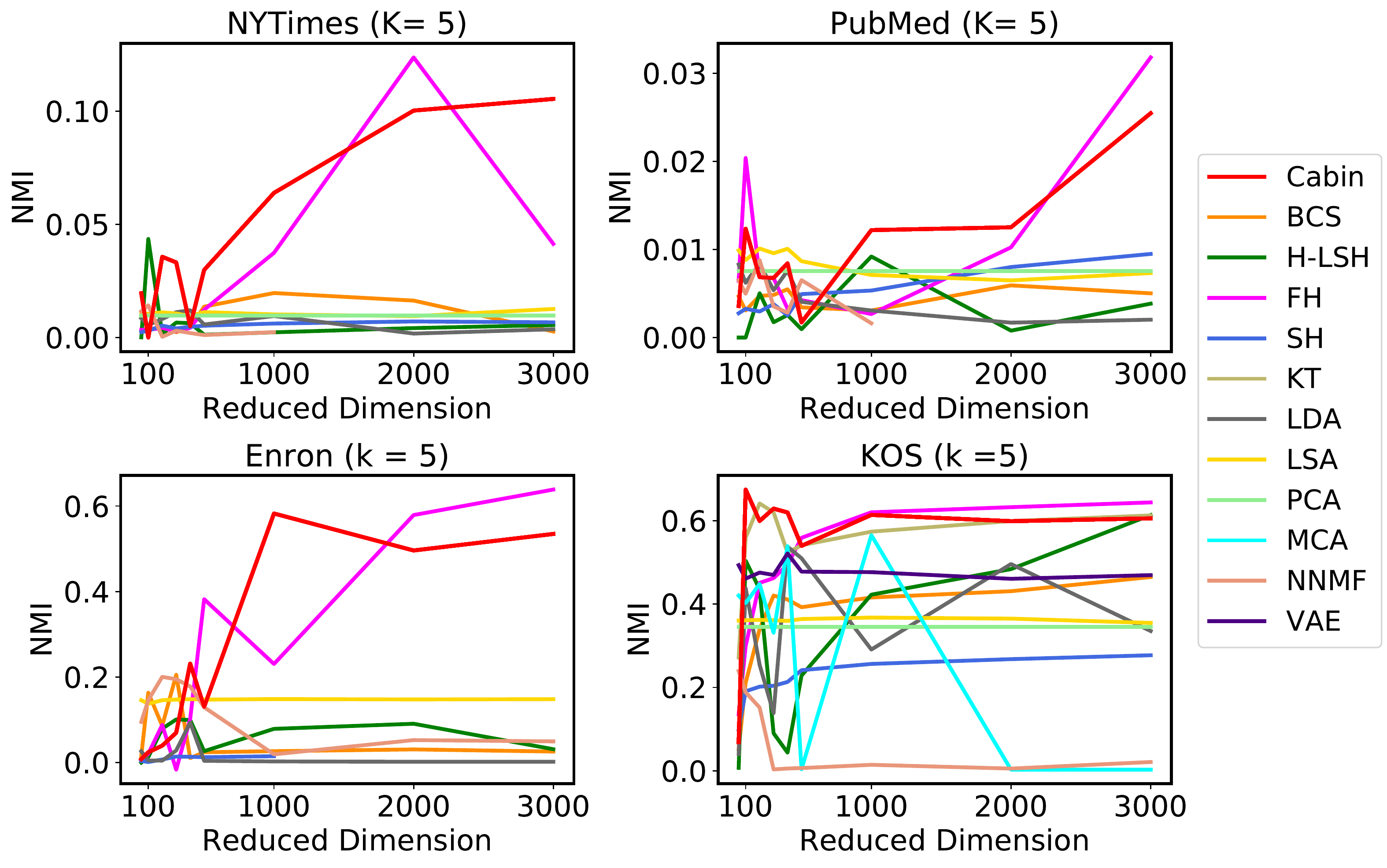}
\caption{ \hl{Comparing the quality of clusters on the compressed datasets using \textit{Normalised Mutual Information (NMI)} metric. The NMI takes a value between $0$ and $1$. A higher value indicates a better performance. Observe that the performance of Cabin is among the top few at 1000 or more embedding dimensions.}}
\label{fig:nmi}
\end{figure*}

\begin{figure*}[ht]
\centering
    \includegraphics[width=0.8\linewidth]{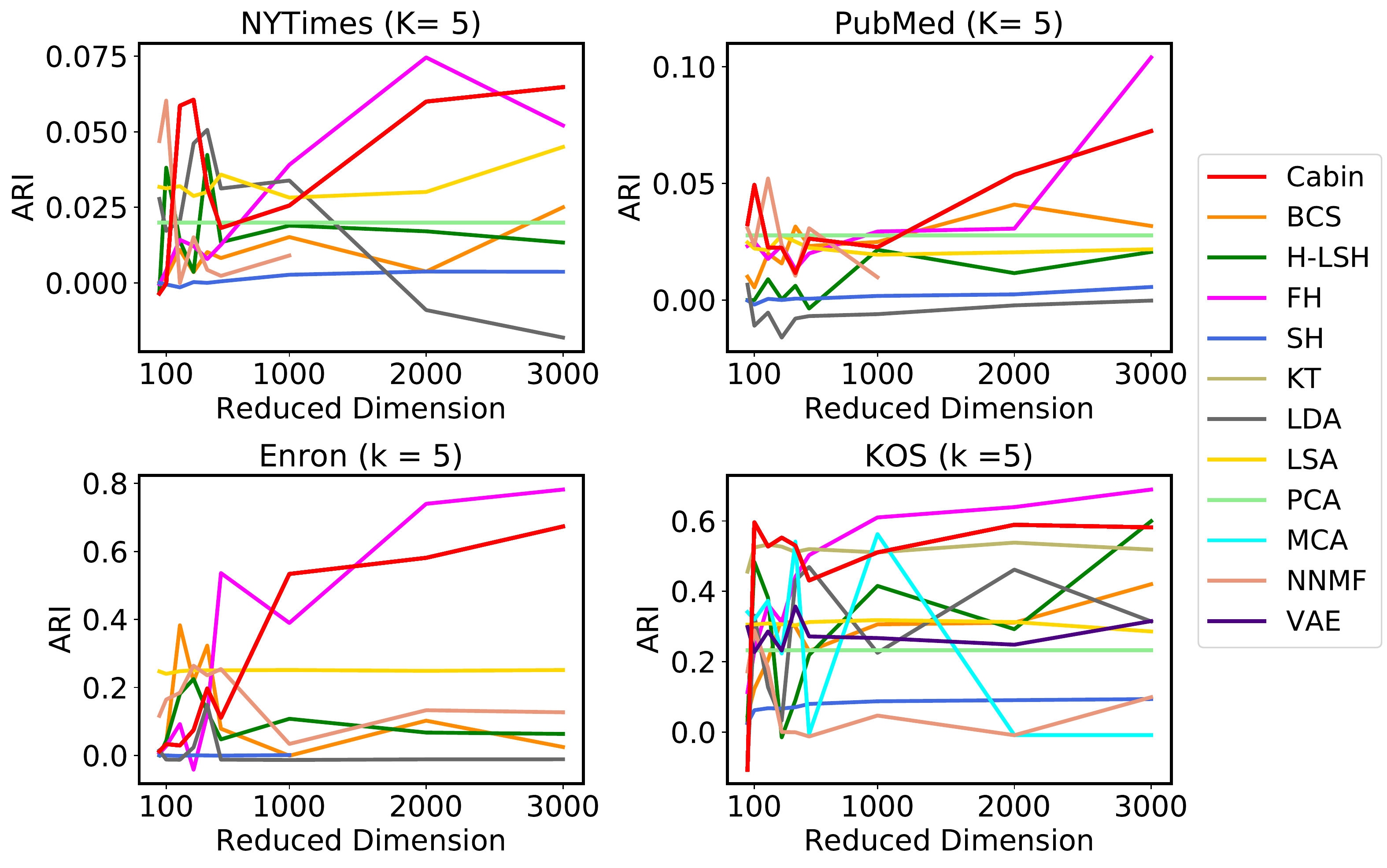}
\caption{ \hl{Comparing the quality of clusters on the compressed datasets using \textit{Adjusted Rand Index  (ARI)} metric. The ARI takes a value between $-1$ and $1$. A higher value indicates a better performance. Observe that  the performance of Cabin is among the top few at 1000 or more embedding dimensions.}}
\label{fig:ari}
\end{figure*}

\begin{figure*}[ht]
\includegraphics[width=0.98\linewidth]{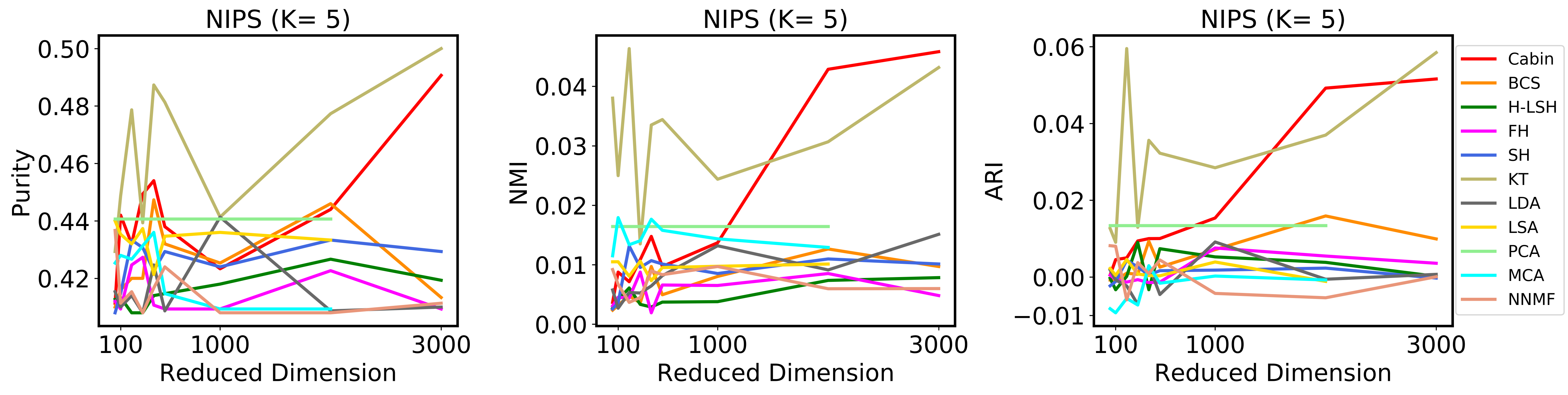}
\caption{ \hl{Comparing the quality of clusters on the compressed NIPS datasets using \textit{purity-index, ARI, NMI} evaluation metrics. Observe that the performance of Cabin is among the top two on all metrics beyond 2000 dimensions.
}}
\label{fig:nips}
\end{figure*}

\begin{figure}[b]
\centering
    \includegraphics[width=0.6\linewidth]{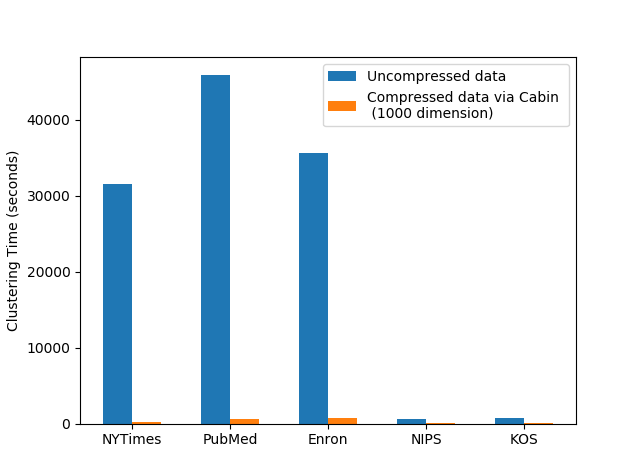}
      \begin{tabular}{c|c|c|c|c|r}
    \hline
    Brain Cell &  NYTimes     &  PubMed        &    Enron        &  {NIPS}&  KOS         \\
    \hline
    {OOM}      & $112.3\times$ &  $70.7\times$  &  $45.1\times$ &  {$10.48\times$}  &   $6.5\times$  \\
 \hline
  \end{tabular}
\caption{{ Comparison of the running time of clustering on the uncompressed dataset with the $1000$ dimensional sketch obtained via \tsketch. Clustering on the full-dimension Brain Cell dataset could not be run on the server that we used for experiments.
}}
\label{fig:clustering_speed_up_bar_plot}
\end{figure}

We first generated the ground-truth clustering on the original dataset using the classical $k$-mode algorithm~\cite{kmode} for several values of $k$. The $k$-mode algorithm is analogous to $k$-means but for Hamming distance. We then performed dimensionality reduction using the available techniques for multiple values of reduced dimensions. Note that several baselines such as LDA, LSA, PCA, MCA, and NNMF  generate real-valued sketches, therefore, instead of $k$-mode, we ran $k$-means (using $k$-means++ sampling distribution~\cite{kmeanspp}) to generate a clustering. We compare the times taken for clustering and the quality of clustering on the reduced dimensions with the ground truth clustering results. For evaluation the quality of clustering we employed standard metrics such as \textit{purity index},   {\textit{adjusted rand index  (ARI)}, and \textit{normalised mutual information (NMI)}; we have explained all these metrics in Subsection~\ref{subsec:clustering_metrics}}.
 
We used the same random seed for all the baselines to ensure that all of them are initialised with the same set of cluster centres in their first iteration. Thus their respective final clustering results \hl{ought to} depend only on the quality of sketches, and not on the randomisation involved in the $k$ initial cluster centres. 

 \noindent\textbf{Observations:} 
  The comparisons of clustering quality on NYTimes, Enron, PubMed, and KOS are presented in Figures~\ref{fig:purity},~\ref{fig:nmi}, and ~\ref{fig:ari} for \textit{purity-index}, \textit{NMI}, and \textit{ARI} evaluation metrics, respectively. The same on the NIPS dataset in presented in Figure~\ref{fig:nips}.

 For all three evaluation metrics (\textit{purity-index, ARI} and {\tt NMI}),  the clustering results achieved using our method is quite high; it consistently remains one of the top \hl{two} approaches across all the bag-of-words datasets and \hl{for all but very small dimensions. It should be noted that, theoretically, the embedding dimenssion should be above a certain minimum that depends upon the sparsity of a dataset. Nevertheless, we ran the experiments at very low dimensions, and were pleasanly surprised to see good clustering scores, e.g., for the KOS dataset even at a dimension as low as 100.}

\hl{We explained earlier that succinct binary sketches offer significant advantage during data analytic tasks. To illustrate this advantage, we compare the speedups obtained by clustering 1000-dimensional \tsketch sketches {\it vs.} the uncompressed data points belonging to the different datasets. The speedup is presented in 
Figure~\ref{fig:clustering_speed_up_bar_plot} and is significant, e.g., $112.3\times$ on the NYTimes dataset.}


 \begin{figure*}[ht]
\centering
\includegraphics[width=\linewidth]{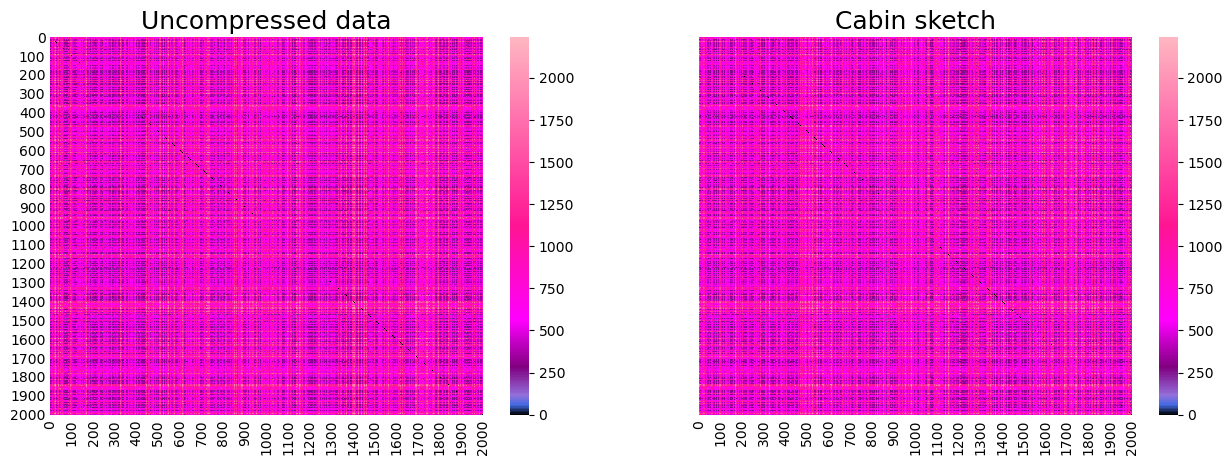}
\caption{Heat-maps for pairwise Hamming distance estimation. The left heat-map  corresponds to pairwise Hamming distance on the full-dimensional Brain Cell dataset, and the right  heat-map corresponds to pairwise Hamming distance estimation from a $1000$ dimensional sketch of Brain Cell dataset obtained via \tsketch. 
}
\label{fig:CAB_heatmap}
\end{figure*}

\begin{figure*}[ht]
\centering
\includegraphics[width=\linewidth]{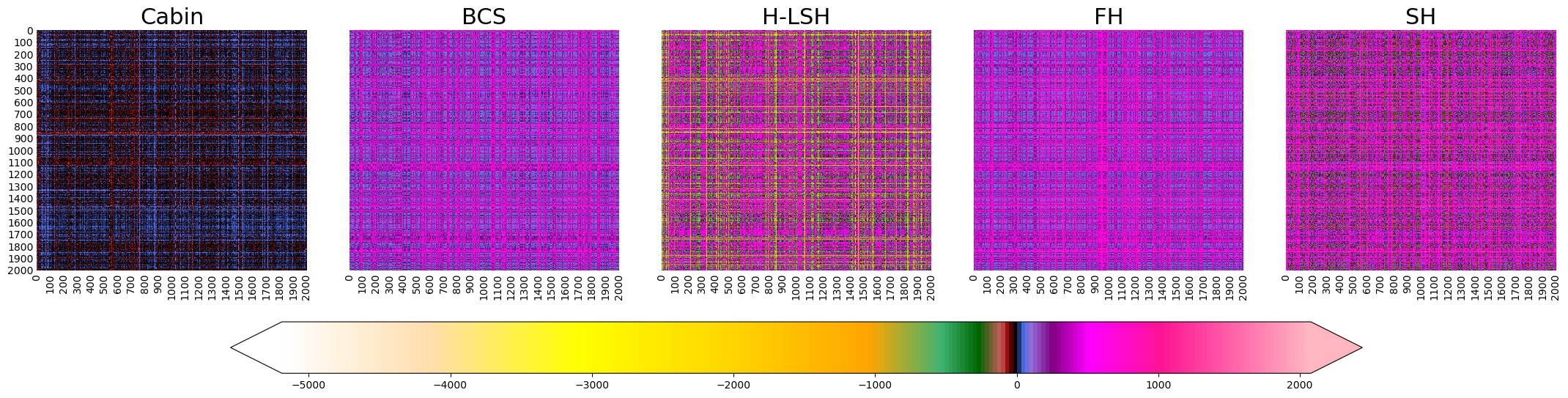}
\caption{Heat-maps on pairwise Hamming error among the baselines. The darker colours correspond to zero error or close to zero error and indicate  better performance. Also, see Table~\ref{tab:hamming_error_heat_map} in which we have compared the MAE as well.}
\label{fig:heat_map_hamming_error}
\end{figure*}

\subsection{Generation of all-pairs similarity matrix}
A common data analysis task is to generate a ``all pairs similarity matrix'', {\it aka.} heatmap (also referred to as pairwise distance/similarity matrix in libraries and tool-kits)~\cite{heat-map-application}. A heatmap refers to an $N \times N$ matrix, where $N$ denotes the number of data points, whose $(i,j)$-th entry stores the (dis)similarity between the $i$-th and the $j$-th data points. Generating a heatmap of a high-dimensional categorical dataset is tricky since a significant amount of time is spent on computing the pairwise Hamming distances. Our earlier experiments show that \hamest is able to estimate Hamming distances from low-dimensional binary sketches with high accuracy, so we conducted an experiment to evaluate how well and how fast our method can generate a heatmap.

\hl{For this experiment} we took $2000$ data points from the Brain cell data set and generated the pairwise Hamming distance matrix (left image in Figure~\ref{fig:CAB_heatmap}). Then we reduced the dimension of those points to $1000$ using \tsketch and other baselines that output discrete sketches, namely, BCS, Hamming-LSH, SimHash, Feature Hashing (KT was not included due to OOM error). Heatmaps were generated from the respective sketches (the heatmap from \tsketch is shown as the right image in Figure~\ref{fig:CAB_heatmap}).

\begin{table}[!ht]
    
\caption{{ Comparing Mean Absolute Hamming error (MAE) in  Heat-maps. 
} }\label{tab:hamming_error_heat_map}
 \begin{center}
  \begin{tabular}{lccccr}
    \hline
        Mean abs. Ham.                  &    Cabin    &     BCS     &    H-LSH       &    FH        &   SH        \\
    \hline
    error (MAE)&    $23.86$  &  $281.19$   &  $505.23$    &  $351.01$    &   $368.32$  \\

 \hline
  \end{tabular}
  \end{center}
\end{table}
\noindent\textbf{Observations:} A quick glance shows that both the heatmaps in Figure~\ref{fig:CAB_heatmap} are visually quite similar and it appears that the heatmap from \tsketch sketches can very well substitute for the latter. \hl{The visual similarity may not be very convincing, and further, to compare against the other methods, we generated a heatmap of the Hamming errors for all the appropriate methods; the heatmaps are displayed in Figure~\ref{fig:heat_map_hamming_error}. We also tabulated the mean (absolute) Hamming errors (MAE) in Table~\ref{tab:hamming_error_heat_map}.} 
\hl{The MAE of our solution is less than $1/10$-{\it th} of those arising out of the baseline approaches and the heatmap of the Hamming errors of \tsketch sketches is markedly superior compared to the sketches of the other methods.}
It is clearly evident that the heatmap from \tsketch sketches is a close approximation to the actual heatmap and the best, by far, among all the other alternatives.


\hl{Efficient algorithms for generating an all-pairs similarity matrix is an emerging area~\cite{spaen-thesis}.} In this context, we found it quite surprising that \tsketch is able to  compress the Brain cell dataset from $1306127$ dimensions to $1000$ dimensions and is still able to accurately estimate the pairwise Hamming distances. Heatmap generation on the compressed data takes about \textbf{$\mathbf{570~\mu}$}sec compared to about $\mathbf{78}$ msec on the uncompressed version for each entry of the matrix, leading to roughly $\mathbf{136\times}$ speedup.

\paragraph{Errors during dimensionality reduction:}\label{sec:appendix_oom_error}
We noticed that several baseline methods give an \textit{out-of-memory (OOM)} error or their reported running time is quite high, especially on high dimensional datasets. 
 For example,  {VAE reported \textit{OOM} error on all datasets except KOS}, 
KT threw \textit{OOM} error on NYTimes, PubMed, Brain Cell, and on Enron, it didn’t stop even after $20$ hrs. MCA also reported \textit{OOM} error for NYTimes, PubMed, and Brain Cell datasets. Further, the dimensionality reduction time for NNMF was quite high; on NYTimes, it took around $20$ hrs to reduce to $3000$ dimensions, and on PubMed and Brain cell dataset, it didn’t stop even after $20$ hrs. We, therefore, could not perform dimensionality reduction with some of the algorithms for all dimensions.

\section{Conclusion}\label{sec:conclusion}
In this work, we propose an efficient dimensionality reduction algorithm for sparse categorical data that takes high-dimensional categorical vectors as input and outputs low-dimensional binary vectors. We also present an algorithm to closely approximate the Hamming distance between the original data from the low-dimensional vectors. With the help of theoretical analysis and extensive experiments, we establish that our approach can be used to reduce the dimension of high-dimensional sparse datasets before sending them for data analysis to speed up the tasks without hurting their results.

\hl{The novelty of our method is that it strives to be task agnostic. Based on the observations in the RMSE and MAE experiments, we can recommend that a low-dimensional \tsketch-compressed dataset can safely be used in the place of a high-dimensional categorical dataset for any task that relies on the Hamming distances  between the data points. Our experiments revealed that there could be alternatives that excel at some particular task (e.g., using the Kendall-tau rank correlation compression algorithm for clustering the ``NIPS full paper'' dataset), but if a go-to approach is required, then there is really no other alternative. To top it off, our solution comes with theoretical guarantees that explain when and why it can be really useful.}

There are several interesting directions along which our approach can be further improved. Even though we use completely random maps ($\pi$ and $\psi$) during hashing, it may be possible to design better hash functions with some knowledge of the distribution of sparsity beyond a simple upper bound on it that \tsketch uses. There is also the theoretical question of using the best-possible and least-resource intensive hash functions, e.g., whether pairwise-uniform hash functions are suitable for \tsketch. Even though our proposed solution is entirely unsupervised, we appreciate the benefits of supervised learning and we think that it may be possible to ``learn'' the right hash functions from training data while retaining the underlying properties of \tsketch and \hamest.

\bibliographystyle{plain}


\end{document}